\documentclass{article}

\usepackage{arxiv}
\usepackage{tabularx}

\usepackage[utf8]{inputenc} 
\usepackage[T1]{fontenc}    
\usepackage{hyperref}       
\usepackage{url}            
\usepackage{booktabs}       
\usepackage{amsfonts}       
\usepackage{nicefrac}       
\usepackage{microtype}      
\usepackage{lipsum}
\usepackage{graphicx}
\usepackage{wrapfig}
\graphicspath{ {./images/} }
\usepackage{gensymb}

\title{A Survey on Proactive Customer Care: Enabling Science and Steps to Realize it}

\author{
Viswanath Ganapathy \\
  Advanced AI, America Research Lab, LG Electronics\\
  5150 Great America Pkwy\\
  Santa Clara, CA 95055 \\
  \texttt{gviswa@gmail.com} \\
   \And
 Sauptik Dhar \\
  Advanced AI, America Research Lab, LG Electronics\\
  5150 Great America Pkwy\\
  Santa Clara, CA 95055 \\
  \texttt{sauptik.dhar@gmail.com} \\
   \And
 Olimpiya Saha \\
  Advanced AI, America Research Lab, LG Electronics\\
  5150 Great America Pkwy\\
  Santa Clara, CA 95055 \\
  \texttt{osaha@unomaha.edu} \\
   \And 
 Pelin Kurt Garberson \\
  Advanced AI, America Research Lab, LG Electronics\\
  5150 Great America Pkwy\\
  Santa Clara, CA 95055 \\
  \texttt{pelin.kurt.4d@gmail.com} \\
  \And
  Javad Heydari \\
  Advanced AI, America Research Lab, LG Electronics\\
  5150 Great America Pkwy\\
  Santa Clara, CA 95055 \\
  \texttt{khormizi@gmail.com} \\
  \And
 Mohak Shah \\
  Advanced AI, America Research Lab, LG Electronics\\
  5150 Great America Pkwy\\
  Santa Clara, CA 95055 \\
  \texttt{mohak.shah@lge.com} \\
}

\begin{document}
\maketitle
\begin{abstract}
In recent times, advances in artificial intelligence (AI) and IoT have enabled seamless and viable
maintenance of appliances in home and building environments. Several studies have shown that AI has the
potential to provide personalized customer support which could predict and avoid errors more reliably
than ever before. In this paper, we have analyzed the various building blocks needed to enable a successful AI-driven predictive maintenance use-case. Unlike, existing surveys which mostly provide a deep dive into
the recent AI algorithms for Predictive Maintenance (PdM), our survey provides the complete view\mbox{;} starting from business impact to recent technology advancements in algorithms as well as systems research and model deployment. Furthermore, we provide exemplar use-cases on predictive maintenance of appliances using publicly available data sets. Our survey can serve as a template needed to design a successful predictive maintenance use-case. 
Finally, we touch upon existing public data sources and provide a step-wise breakdown of an AI-driven proactive customer care (PCC) use-case, starting from generic anomaly detection to fault prediction and finally root-cause analysis. We highlight how such a step-wise approach can be advantageous for accurate model building and helpful for gaining insights into predictive maintenance of electromechanical appliances.
\end{abstract}

\section{Introduction}
\begin{wrapfigure}{L}{0.45\textwidth}
\centering
\includegraphics[width=0.4\textwidth]{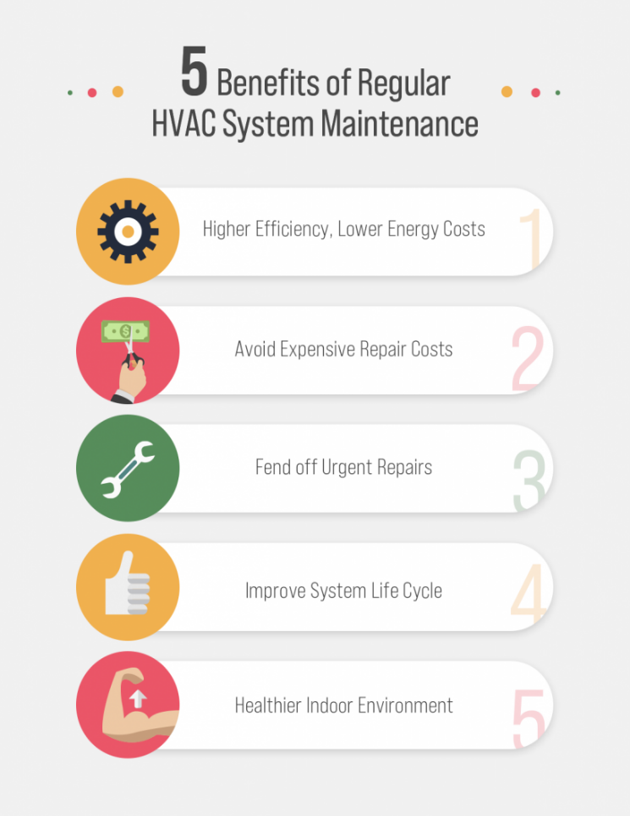}
\caption{Benefits of proper device maintenance \cite{LG}}\label{ACbenefit}
\end{wrapfigure}
Maintenance as a crucial activity in industry, with its significant impact on costs and reliability, is immensely influential to a company’s ability to be competitive in low price, high quality and performance. Any unplanned downtime of machinery equipment or devices would degrade or interrupt a company’s core business; potentially resulting in significant penalties and immeasurable reputation loss.  For instance, a 49 minutes of downtime cost Amazon $4$ million in lost sales in 2013     \cite{Forbes082013}. According to a market study by the Ponemon Institute \cite{Ponem2019}, organizations in average lose $138,000$ per hour due to data centre downtime. Production issues at Volkswagen in 2016 lead to dramatic losses in sales of up to 400 million Euros per week. GE report \cite{GEOil2016} highlights the losses in oil and gas sector due to unplanned downtime. These examples show the huge financial impact of a working production facility for companies. It was also reported that the Operation and Maintenance (O\&M) costs for offshore wind turbines account for 20\% to 35\% of the total revenues of the generated electricity and maintenance expenditure in oil and gas industry costs ranging from 15\% to 70\% of total production cost. In the data-driven domains of Industry 4.0 and Industrial IoT with intelligent, connected machines, approaches beyond static maintenance schedule is feasible.  Therefore, it is critical for companies to develop a well-implemented and efficient maintenance strategy to prevent unexpected outages and improve overall reliability with reduced operating costs. 

In the context of household  appliances, the operational conditions could vary and  could result in abnormal operation. These abnormal conditions can be attributed to different causes that identify an appliance as anomalous. The power consumption pattern of an anomalous appliance deviates from its expected normal behavior.  Anomalous appliances can result in inefficient energy usage, reduced operation performance, and impede safety. In the recent times, it was reported that abnormal usage of refrigerator resulted in a massive fire \cite{Grenfell2017}. Accordingly, household appliances' proper maintenance is needed by customers to reduce the energy costs, ensure safety and for system operators to enable energy efficiency improvements. SAC/Chiller business unit of a leading manufacturer emphasises the need for proper maintenance and identifies 5 major benefits for appliance (HVAC) maintenance  (see Fig. \ref{ACbenefit}). In addition, they also provide weekly, monthly and yearly maintenance schedules for customers through their online articles \cite{LG}.


In addition to maintenance schedule guidelines to the customers, most OEMs (Original Equipment Manufacturers) provide 3 types of maintenance support viz.,  Reactive Maintenance (RM), Periodic Maintenance (PM) and Predictive Maintenance (PdM). 
RM is only executed to restore the operating state of the equipment after failure occurs, and thus tends to cause serious lag and results in high reactive repair costs. PM is carried out according to a planned schedule based on time or process iterations to prevent breakdown, and thus may perform unnecessary maintenance and result in high prevention costs. In order to achieve the best trade-off between the two, PdM is performed based on an online estimate of the “health” and can achieve timely pre-failure interventions. PdM allows the maintenance frequency to be as low as possible to prevent unplanned RM, without incurring costs associated with doing too much PM. PdM typically involves condition monitoring, fault diagnosis, fault prognosis, and maintenance plans. The evolution of modern techniques (e.g., Internet of things, sensing technology, artificial intelligence, etc.) has resulted to significant gains through PdM compared to RM or PM. In this manuscript we use PdM and Predictive Customer Care (PCC) interchangeably in the context of alerting end customers the state of an appliance.

Unlike existing surveys, which mostly provide a deep dive into the recent AI algorithms for Predictive Maintenance, this survey provides the complete view; starting from business impact, to the recent technology advancements in algorithms as well as systems research and model deployment. Our survey highlights the advantages of PCC, the underlying developments in AI and engineering advances in Big Data Analytics that make it a reality. We also discuss a subset of AI algorithms that can be effectively leveraged for predicting the condition of electromechanical systems as well as for identifying the most likely cause influencing the condition. The real world engineering components which enable Big Data and it's impact on realizing PCC are succinctly included in this manuscript.  PCC envisages the ability to pre-empt and solve errors faster than ever before on every connected appliance. This manuscript includes a set of concrete steps to realize our PCC vision. \\

\noindent \textbf{Organization of the manuscript} : In this manuscript we first discuss the different maintenance services provided by most OEMs (Section \ref{sec:cat_maintenance}). We identify the significant benefits obtained through Predictive Maintenance as compared to Reactive or Periodic Maintenance, and briefly touch upon some important business and technology aspects of PdM (Section \ref{subsect_pred_maintenance}). Section \ref{sec_technology} delves deeper into the technology advancements pertaining to PdM, and the other aspects like AI Infrastructure and deployment needed for a successful PdM use-case. Next, we provide concrete instances of machine learning model development for PdM use-cases in section \ref{sec_case_studies}. In section \ref{external_sources} we provide a list of model development platforms and external data / code sources which forms the starting point for any PdM use-case. We contextualize this manuscript from the perspective of PdM for electromechanical appliances. Further, we discuss the lessons learned from our analysis of open source data-sets for  PdM in section \ref{sec_lessons_learnt}. Section \ref{conclusions} provides the conclusions.

\section{Categories of Maintenance} \label{sec:cat_maintenance}
 Maintenance approaches fall into one of the three categories. 
 
\subsection{Reactive Maintenance (RM)}
Under reactive maintenance (RM) approach the maintenance action for repairing equipment is performed only when the equipment breaks down or when the equipment has run to the point of failure. The RM approach usually leads to equipment failure and results in downtime.

\subsection{Preventive Maintenance (PM)}
Preventive Maintenance (PM), also referred to as planned maintenance, schedules regular maintenance activities on specific equipment to lessen the likelihood of failures.The maintenance is executed even when the machine is still working and under normal operation so that the unexpected breakdowns with the associated downtime and costs would be avoided. Almost all PM management programs are time-driven. It is assumed that the failure behavior (characteristic) of the equipment is predictable and it follows the well known bathtub curves. The bathtub curve indicates that new equipment would experience a high probability of failure due to installation problems during the first few weeks of operation. After this break-in period, the failure rate becomes relatively low for an extended period. After the normal life period, the probability of failure increases dramatically with elapsed time. The general process of PM can be presented in two steps: 1) The first step is to statistically investigate the failure characteristics of the equipment based on the set of time series data collected. 2) The second step is to decide the optimal maintenance policies that maximize the system reliability/availability and safety performance at the lowest maintenance costs. PM could reduce the repair costs and unplanned downtime, but might result in unnecessary repairs or catastrophic failures. Determining when a piece of equipment will enter the wear out phase is based on the theoretical rate of failure instead of actual status on the condition of the specific equipment. This often results in costly and completely unnecessary maintenance taking place before there is an actual problem or after the potentially catastrophic damage has begun. Also, this will lead to much more planned downtime and require complicated inventory management. In particular, if the equipment fails before the estimated ”ware out” time, it must be repaired using RM techniques. Existing analysis has shown that the maintenance cost of repairs made in a reactive mode (i.e., after failure) is normally three times greater than that made on a scheduled basis.

\subsection{Predictive Maintenance (PdM) a.k.a Proactive Customer Care} \label{subsect_pred_maintenance}
Predictive Maintenance (PdM), also known as condition-based maintenance (CBM) or Proactive Customer Care (PCC), aims to predict when the  equipment is likely to fail and decide which maintenance activity should be performed such that a good trade-off between maintenance frequency and cost can be achieved. The principal of PdM is to use the actual operating condition of systems and components to optimize the O\&M. The PdM approaches leverage physics based models as well as data collected from multiple sources including sensors. Predictive analysis is based on data collected from meters or sensors connected to machines and tools such as, vibration data, thermal images, ultrasonic data, operation availability, etc. The predictive model processes the information through predictive algorithms, discovers trends and identifies when equipment will need to be repaired or retired. Rather than running a piece of equipment or a component to failure, or replacing it when it still has useful life. PdM helps companies to optimize their scheduled maintenance only when absolutely necessary. With PdM, planned and unplanned downtime, high maintenance costs, unnecessary inventory and unnecessary maintenance activities on working equipment can be significantly decreased. In fact, a study by EPRI shows Predictive Maintenance care provides, (see \cite{amiral}) 
\begin{itemize}
\item[--] 30\% cost reduction over periodic maintenance,
\item[--] 50\% cost reduction over reactive maintenance.
\end{itemize}
The advantages of PdM has also been confirmed by US DOE which reports (see \cite{usdoe}),
\begin{itemize}
    \item[--] 	Return on investment: 10 times 
 \item[--] Reduction in maintenance costs: 25\% to 30\%  
\item[--] Elimination of breakdowns: 70\% to 75\% 
\item[--] Reduction in downtime: 35\% to 45\% 
\item[--] Increase in production: 20\% to 25\% 
\end{itemize}

\begin{wrapfigure}{R}{0.60\textwidth}
\centering
\includegraphics[width=0.55\textwidth]{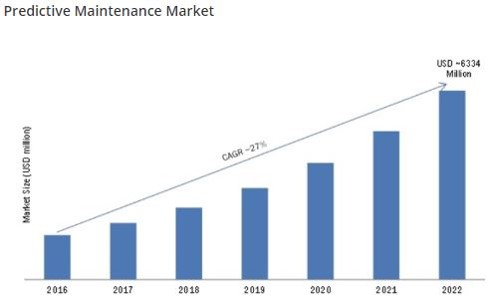}
\caption{Growth of PdM Market}\label{fig_Pdm_Market}
\end{wrapfigure}

In fact, a report from marketresearchfuture.com (see Fig. \ref{fig_Pdm_Market}) estimates the growth of PdM market to upto  ~ 6K Million USD by 2022.  Although the benefits of PdM far outweighs its shortcomings. One major limitation of PdM is that, compared to RM or PM, the cost of the condition monitoring devices (e.g., sensors) needed for PdM is often higher. Also, building effective / accurate PdM systems is non-trivial, due to the complexities involving data collection, data analysis, decision making etc.

\subsubsection{AI for Predictive Maintenance}

Lately there has seen a wide adoption of AI to handle the intricacies behind building an effective PdM system. Compared to the traditional PdM approaches adopting rule based models, the more advanced AI driven models tend to be more robust and also scale well with system complexity \cite{sparkcognition2020}. Of course, such AI driven approaches require significant amount of data. The data required for PdM is typically the sensor measurements, warranty logs, maintenance logs, engineering drawing etc. Machine learning algorithms analyze these `big' data sets and identify useful patterns (like novelties or anomalies) to build an effective PdM system. In addition, with the recent advancements in causal inference, advanced techniques can be adopted to learn the underlying  causal relationships between the variables dictating the overall behaviour of the underlying physical model. This greatly helps towards building accurate predictive diagnostic models.


In addition to building predictive models, it is beneficial to provide user guidelines for effectively managing the electromechanical devices based on PdM analysis. This is known as prescriptive maintenance, and is typically considered a natural extension of PdM analysis. Prescriptive maintenance builds on the insights from predictive models to help determine why an event happened and lists the set of actions to influence the best outcome. The Gartner analytics ascendancy model \cite{gartner2019} suggests that prescriptive techniques would bring significant value even though it is challenging to design and deploy.


\subsubsection{Business impact of AI based Predictive Maintenance}

 Forecasting failures in manufacturing process as well as equipments can significantly minimize losses due to unplanned down times. It was recently reported that unplanned downtime costs the industry an estimated 50 billion USD annually \cite{cloudera2020}. AI driven PdM has shown promise across different industries minimize downtime and reduce losses \cite{cloudera2020,sparkcognition2020}. In the Airline sector PdM approaches have significantly improved in flight safety, reduced maintenance time of air-crafts as well as significantly improved supply chain management \cite{c3ai2020}. A primary money-saving solution is using machine learning to predict when aircraft will need maintenance to avoid malfunctions and extend the lifetime of the aerial vehicles. In the power distribution sector PdM based approaches have avoided major shutdowns and enabled significant cost savings. In the oil and gas sector AI driven approaches have enabled significant increase in production.

For home appliance sector it is predicted that the incorporation of AI driven PdM along with prescriptive maintenance would reduce the number of technician visits as well as increase the operational efficiency of every appliance. Detecting anomalous behavior of appliances that are working all the time, e.g., refrigerators, is also expected to bring down energy usage. In the absence of PdM, users are unaware of the variation in energy usage due to anomalies in appliances such as HVACs, refrigerators, air purifiers etc. Further, usage suggestions as well as maintenance suggestions provided by PdM can minimize unexpected breakdowns as well as repairs. This results in improved customer experience.

The data analytics engine built for PdM can be employed for improved product design and engineering. The PdM data analytics can also be leveraged for several other value added services including competitive pricing for insurance \cite{ibm2016}.



\section{Recent Technologies for Predictive Maintenance (PdM)} \label{sec_technology}
As discussed in the previous section \ref{subsect_pred_maintenance}, predictive maintenance (PdM) enables cost effective solutions with gains  of up to $\sim 50 \%$  compared to reactive, and up to $\sim 30 \%$  compared to preventive maintenance \cite{amiral}. So, it is of utmost importance to have a deeper understanding on the recent trends in predictive maintenance. This section covers the recent technology advancements in predictive maintenance. Note that, at the core of any predictive maintenance use-case is correct understanding of the business use-case's return on investment (ROI) and appropriate design of a use-case specific evaluation metric. Traditional machine learning evaluation metrics like, (balanced) accuracy, area under curve (AUC) etc., may not yield optimal business value for such use-cases. There are several research towards better capturing the ROIs using modified evaluation metrics. However, such metrics are use-case specific and heavily depends on the company's expected target returns. Rather, we keep this section generic and analyze the algorithmic developments in PdM casted as a traditional ML problem using balanced accuracy or AUC as evaluation metric.          

\subsection{Anomaly Detection for PdM} \label{sec:AnomalyPDM}
A widely used approach is to cast PdM as an Anomaly detection problem. The main idea is to estimate a model to identify device / equipment errors (anomalies)  within a prediction horizon. There are several technology advancements pertaining to PdM using anomaly detection. We categorize these advancements on the basis of traditional vs. modern deep learning based technologies next. 

\subsubsection{Traditional ML based Anomaly Detection for PdM}
There have been significant research on anomaly detection for PdM. There are several surveys providing an in-depth analysis of the existing technologies \cite{chandola2009anomaly,basora2019recent}. In this work we provide a very brief summary of the typical approaches adopted for most existing approaches. Majority of existing research can be broadly categorised into one of the following approaches (see \cite{chandola2009anomaly}).
\begin{enumerate}
    \item \textbf{Parametric Statistical Modeling}: These approaches assume that the data distribution follow an underlying model $f(\Theta, \mathbf{x})$, where $\mathbf{x} -$ input data, $\Theta - $ model parameters. The goal behind these approaches is to estimate the parameters of the data generation process $f(\Theta, \mathbf{x})$. The typical adopted model parameterization include,  gaussian model based, regression model based, mixture of distribution based techniques, etc. 
    \item \textbf{Non-parametric Statistical Modeling}:  These approaches do not assume any parametric form for the underlying data distribution. That is, no apriori structure for the underlying data distribution is assumed. Some typical approaches include, Histogram based, Kernel or Parzen Windows based approach, etc.
    \item \textbf{Classification Based approaches}: These approaches model the anomaly deteion problem as a single class or extreme rare class problem. Some widely adopted approaches include, Support Vector Machine, Neural Networks, Bayesian Networks, etc. 
    \item \textbf{Spectral}: Spectral based techniques involve analyzing the data variance in some transformed space. Typical examples include, Principal Component Analysis, Linear Discriminant analysis, etc. 
\end{enumerate}

A table of the representative approaches is reproduced from \cite{chandola2009anomaly} and  provided in \ref{tab:ML}.

\begin{table}[h]
\centering
\caption{Traditional ML approaches for Anomaly Detection based PdM. See \cite{chandola2009anomaly} for a more exhaustive coverage.} 
\label{tab:ML}

\begin{tabular}{|c|c|}  
\noalign{
\hrule height 2pt
}
\textbf{Approach} & \textbf{Existing Literature} \\
\noalign{
\hrule height 2pt
}
Parametric Statistical Modeling & \cite{keogh2007finding,guttormsson1999elliptical,keogh1997probabilistic,keogh2002finding,ruotolo1997statistical} \\
Non Parametric Modeling & \cite{manson2002identifying,manson2001long,manson2000long,desforges1998applications} \\
Classification Based & \cite{bishop1994novelty,campbell2000linear,diaz2002residual,harris1993neural,jakubek2002fault,king2002use,li2002improving,petsche1996neural,streifel1996detection,whitehead1995function,brotherton1998classification,nairac1999system,surace1997novelty,surace1998novelty,worden1997structural}\\
Spectral & \cite{fujimaki2005approach,parra1996statistical}\\
\noalign{
\hrule height 2pt
}
\hline
\end{tabular}
\end{table}

\subsubsection{Deep Learning based Anomaly Detection for PdM} \label{sec:DLAnomalyPDM}
A more recent line of research adopt deep learning architectures for anomaly detection. These approaches learn feature representations or anomaly scores using deep neural networks and typically yield significant performance improvements compared to their conventional counterparts. Most of these approaches predominantly adopts a classification based approach discussed in Section \ref{sec:AnomalyPDM}. The main idea is to build a parametric model using the normal samples (or a small proportion of defective samples for rare class problems) to estimate the support of the normal data distribution. Using this model we construct a decision rule (or a score) to identify if a test sample is anomalous/defective or not. There are a gamut of modern deep-learning methods that adopts this approach. Some of the notable recent approaches providing SOTA performance in benchmark data sets like Table \ref{tab:VisInsp} include,
\begin{enumerate}
    \item AnoGAN (and variants) \cite{schlegl2017unsupervised, schlegl2019f}: These approaches adopt a GAN based approach where the discriminator is trained only using the normal class samples.
    \item Auto Encoder based \cite{bergmann2018improving,wang2004image,kingma2013auto}  : Adopts an auto encoder to model the normal data distribution. During detection the reconstructed anomalous samples do not match with the original samples (pixel wise comparison). This dissimilarity score is used to identify anomalies. A variant on this idea using a conditional variational autoencoder~\cite{bib:ConditionalVAE} learns the expected probability distribution of each feature in the encoder latent space to better identify true outliers.
    \item Modified Loss (typically margin based): These approaches do not modify the underlying deep-architectures (CNN or LSTM etc). Rather, they target to modify the loss function. Some of the notable approaches include, Deep SVDD \cite{ruff2018deep} , DROCC \cite{goyal2020drocc}, DAGMM \cite{zong2018deep}, GOAD \cite{bergman2020classification}, Outlier Exposure \cite{dhar2014analysis,hendrycks2018deep,ruff2021unifying}, DOC$^3$ \cite{dhar2021doc3}. 
    \item Time series anomaly detection with generative adversarial networks~\cite{bib:TadGAN} : trains LSTM models to encode and then generate new time series data that are compared with the orignal time series using ``critics" to classify anomalies. 
    \item Graph transformer and convolution networks~\cite{bib:GraphTransformer} : learns a graph representation of the features where the connections represent the relationships between the features, and further, uses a combination of graph convolutions and a transformer to make most efficient use of time series information. This approach may be advised if multiple sensors with physical relationships to one another are recording data.
\end{enumerate}
A more detailed survey on deep learning based anomaly detection is available in \cite{chalapathy2019deep}.





\noindent \textbf{Visual Inspection for PdM}: 
Most of the above approaches cover recent advancements in deep-learning methods for anomaly detection based PdM. One particular application that has recently gained huge prominence is Visual Inspection for PdM. Visual inspection can be considered a "sub-task" within the entire PdM pipeline and involves detecting product faults in a manufacturing/production line using visual (image, video) modalities. A representative subset of these recent research with the respective data sets is provided in \ref{tab:VisInsp}.

\begin{table}
\centering
\caption{Defect detection using Visual Inspection} 
\label{tab:VisInsp}
\begin{tabular}{|p{3.5cm}|p{3cm}|p{4cm}|}  
\noalign{
\hrule height 2pt
}
\textbf{Dataset} & \textbf{Description} & \textbf{Adopted Approaches} \\ \hline
Tilda Textile Texture \cite{tilda} & The goal is to identify fabric defects from images. & Auto-Encoder based \cite{tian2019autoencoder,han2020fabric}, Extreme-Learning based \cite{liu2019unsupervised}, CNN based  \cite{jing2019automatic}. \\ \hline
DAGM Industrial Optical Inspection \cite{dagm2007} & The goal is to identify synthetically generated  defect patterns. & GAN based \cite{Niu2019}, Segmentation based \cite{bovzivc2020end}, Auto-Encoder based \cite{liu2020one},  U-Net \cite{nvidia}, CNN based  \cite{yu2017fully}. \\  \hline
Neu Surface defect detection \cite{song2013noise} & The goal is to detect surface defects on hot rolled steel strip. &  CNN based \cite{Ren2018,yi2017end,zhou2017classification,gao2020semi}, Resnet \cite{song2020edrnet}, GAN based \cite{gao2020generative}, Auto Encoder \cite{kholief2017detection}\\  \hline
Magnetic Tile Defect detection \cite{huang2020surface} & The goal is to detect defects in magnetic tiles. &  Auto Encoder \cite{liu136semi}, CNN Based \cite{minhas2019anonet,wang2020cpacconv} \\  \hline
KolektorSDD data \cite{Tabernik2019JIM} & The goal is to detect the defects in electrical commutators that were provided and annotated by Kolektor Group. &  Segmentation based \cite{Tabernik2019JIM,bovzivc2020end}, YOLO \cite{hatab2020surface} \\  \hline
MVTec Anomaly Detection \cite{Tabernik2019JIM} & The goal is to detect the defects in various manufacturing products and surfaces. & Normalizing Flow \cite{rudolph2020differnet}, GAN, Autoencoder, CNN \cite{bergmann2019mvtec} \\  \hline
\noalign{
\hrule height 2pt
}
\end{tabular}
\end{table}

\subsubsection{Challenges}
Unlike those problems and tasks on majority, regular, or evident patterns, anomaly detection addresses minority, unpredictable/uncertain, and rare events, leading to some unique complexities below that render general deep learning techniques ineffective.
\begin{itemize}
    \item Unknown: Anomalies are associated with many unknowns, e.g., instances with unknown abrupt behaviors, data structures, and distributions. They remain unknown until it actually occur.
    \item Heterogeneous anomaly classes. Anomalies are irregular, and thus, one class of anomalies may demonstrate completely different abnormal characteristics from another class of anomalies. For example, in video surveillance, the abnormal events robbery, traffic accident sand burglary are visually highly different.
    \item Rarity and class imbalance. Anomalies are typically rare data instances, contrasting to normal instances that often account for an overwhelming proportion of the data. Therefore, it is difficult, if not impossible, to collect a large amount of labeled abnormal instances.This results in the unavailability of large-scale labeled data in most applications. The class imbalance is also due to the fact that misclassification of anomalies is normally much more costly than that of normal instances.
    \item Diverse types of anomaly. Three completely different types of anomaly have been explored.
    \begin{itemize}
        \item Point anomalies are individual instances that are anomalous w.r.t. the majority of other individual instances. e.g., the abnormal health indicators of a patient.
        \item Conditional anomalies or contextual anomalies, also refer to individual anomalous instances but in a specific context, i.e., data instances are anomalous in the specific context, otherwise normal.The contexts can be highly different in real-world applications, e.g., sudden temperature drop/increase in a particular temporal context, or rapid credit card transactions in unusual spatial contexts. For example, a temperature record of -$30$ \degree C during hot seasons can be anomalous however, in the context of cold seasons, this report can occur.
        \item Group anomalies, a.k.a. collective anomalies, are a subset of data instances anomalous as a whole w.r.t. the other data instances; the individual members of the collective anomaly may not be anomalies, e.g., exceptionally dense subgraphs formed by fake accounts in social network are anomalies as a collection, but the individual nodes in those subgraphs can be as normal as real accounts. For instance, a washing-machine program consists of individual events such as rinse, drain, and spin. Although these actions are individually normal, their occurrence in a wrong sequence can lead to a collective anomaly. 
    \end{itemize}
\end{itemize}

\subsection{Fault Prediction} \label{sec:fault}
Another setting typically seen in PdM use-cases is the fault prediction approach. Such problems are typically adopt multiclass settings. In this case the goal is to predict the exact fault type. Although not as prevalently used (compared to anomaly detection); there are a few line of work that explores this setting. These include, \\
\noindent \textbf{Fault prediction using acoustic analysis}:
  Skilled maintenance technicians can diagnose a machine’s condition based on sound. However, it is challenging to identify anomaly by listening to machines if the sound is not periodic. Further, even in if the periodicity has a very low frequency anomaly detection is challenging. 
  This further compounded by the fact that the machine sound tends to be non-stationary.  To solve the issue deep learning based approaches have been proposed in literature \cite{ suefusa2020anomalous, 8501554, purohit2019mimii, 8902941}.
  
  AI for household appliance sound event detection and classification is an emerging area of research for intelligent diagnosis and evaluation of home appliances. An open dataset with normal and abnormal sound of home appliances is available for exploration \cite{10.1145/3297156.3297186}. Each category has more than one background noise file. Noise data annotated in the mode. 

\noindent \textbf{Fault prediction using vibration analysis} \cite{2020lee,8758199, 8445684,9259076} :
PdM in industrial environments by monitoring vibration data using spectral techniques has been an active area of research and real world deployment. The industrial deployments tend to deploy accurate accelerometers for vibration measurement. It is an area of interest to explore the efficacy of PdM using low cost commercial vibration sensors. Autoencoder based approaches have been employed for classifying anomalies in vibration data gathered from ceiling fans and washing machines.


\subsection{Reinforcement Learning} \label{sec:RL}

Deep reinforcement learning (DRL) combines reinforcement learning (RL) with deep learning to solve a wide range of complex decision making tasks. The traditional reinforcement learning algorithms such as Q-learning arrive at the optimal action at every state by solving Bellman's equations. The value of a pair of taking an action at a state is measured by the associated reward. Here, it is important to note that Q-learning does not scale well with the complexity of the environment. Therefore, Deep Q-Network (DQN) algorithm employs deep learning networks as  function approximators.  The deep learning network, trained to approximate the complete Q-table, can well scale with the number of possible state-action pairs. Recently, many research efforts have been devoted to applying DRL to the field of PdM.

DRL has been employed to learn anomaly detection models in situations where a partially labeled dataset is available alongwith a large-scale unlabeled dataset \cite{pang2020deep}. In these situations, a typical approach involves leveraging unsupervised learning techniques or employing supervised learning to fit based on the limited anomaly examples. Here the data does not represent the entire set of anomalies. DRL approach attempts to identify anomalies that lie beyond the partially labeled datasets. The DRL approach learns to  exploit existing label data while exploring new classes of anomalies. In the DRL based approach a reward function is designed using the labeled portion of the dataset and anomalies in the unlabelled part of the dataset. This approach improves detection accuracy for all classes of anomalies in a given dataset. Therefore the DRL based approach is of significant practical benefit in several real world scenarios. 

Policy based approach for anomaly detection is another setting that has been employed for anomaly detection from streaming data \cite{YU2020103919}. The agents update the policy incrementally without any domain-specific assumptions about the underlying dataset. \cite{oh2020sequential} leverages inverse reinforcement learning (IRL) to detect anomalies in sequential data. IRL attempts to determine the decision making agent’s underlying reward function from its behavior data. A reward function incentivizes the agent and hence it represents the preference of the agent. IRL can be employed to estimate the reward function and thereby understand the normal behavior.  Using the learned reward function it is possible to determine whether a new observation from the target agent follows a normal pattern or not.

Further, DRL framework has been employed for the optimal management of the operation and maintenance of power grids \cite{8468674}. Typical smart grids are equipped with prognostics and health management capabilities. DRL agents can  exploit the information gathered from prognostic health management devices to select optimal operations and maintenance actions on the system components in the grid.

\subsection{Remaining Useful Life (RUL)}  \label{sec:RUL}
Remaining useful life analysis is very useful of PdM (see \cite{9089252}). RUL has been employed in clinical research to estimate the time to an event, such as death or recurrence of a condition. RUL leads to a special type of learning task that is distinct from classification and regression. This task is termed survival analysis or time-to-event analysis or reliability analysis. Many machine learning algorithms have been adopted to perform survival analysis. This includes Support Vector Machines, Random Forest and Boosting. Recently, deep learning based approaches \cite{DBLP:journals/corr/abs-1907-05146} have been employed to estimate the RUL.

Techniques for estimating RUL is relevant in the context of electromechanical appliances for several reasons. The maintenance as well as usage plan can be drawn based on RUL analysis. Further, the available operational data from appliances is mostly partially observable. This calls for effective techniques to handle missing data. In fact, multiple promising approaches for survival analysis have been developed for handling data which is only partially observable.

\subsection{Root Cause Analysis} \label{sec:pelin-RCA}

A robust predictive maintenance program is an important add-on to leverage data from IoT devices into services that can provide value to IoT customers. An essential piece of this is not just to have accurate algorithms to predict future failures, but also to be able to explain why the algorithm thinks a failure is about to happen in a manner a technician can understand. This explanation can both help build confidence in IoT customers that they should take predictive maintenance suggestions seriously, and help technicians identify and repair a device if they are called in as a result. 

One can roughly think of causal analyses as identifying either direct causes or root causes. Direct causes are directly measurable features that influence the identification of an error (typically features of the fault detection model), while root causes indicate the underlying reason for the cause, often not directly measurable. For example, if a part fails, the direct cause may be because it had too much vibration, while the root cause may be that a technician did not install the part correctly in the first place. Typically, the underlying root cause is much harder to determine. One recent paper~\cite{bib:rootCausePredMaint} proposes two approaches to identifying direct cause features, one approach using explainable trees, and another based on the $z$-score of the change in feature relative to its median value across other rows in the dataset. A drawback of the first approach is that it forces one to stick to simple tree-based algorithms with their limited performance, while a drawback of the second approach is that it bases its causal determinations on how unusual the values of the features are rather than anything related to what the model thinks. 

Moving from direct causal analyses to root cause analyses can be tricky. One suggested approach is to have a domain expert map out the conditional relationships between possible root causes in a Bayesian Network. A Bayesian Network is a probabilistic graphical model which concisely describes the relationship between many random variables and their conditional independence via an acyclic directed graph. Once the topology of the Bayesian network is laid out by a domain expert, the conditional probabilities can be learnt from existing data, thereby completely describing the system using the notion of joint probabilities. More details on this approach are discussed here~\cite{bib:rootCauseWebPage}, and a number of other suggested approaches are discussed here~\cite{bib:rootCausePaper}.

\subsection{AI Infrastructure for PdM Pipeline}

\begin{figure}
\centering
\includegraphics[width=\textwidth]{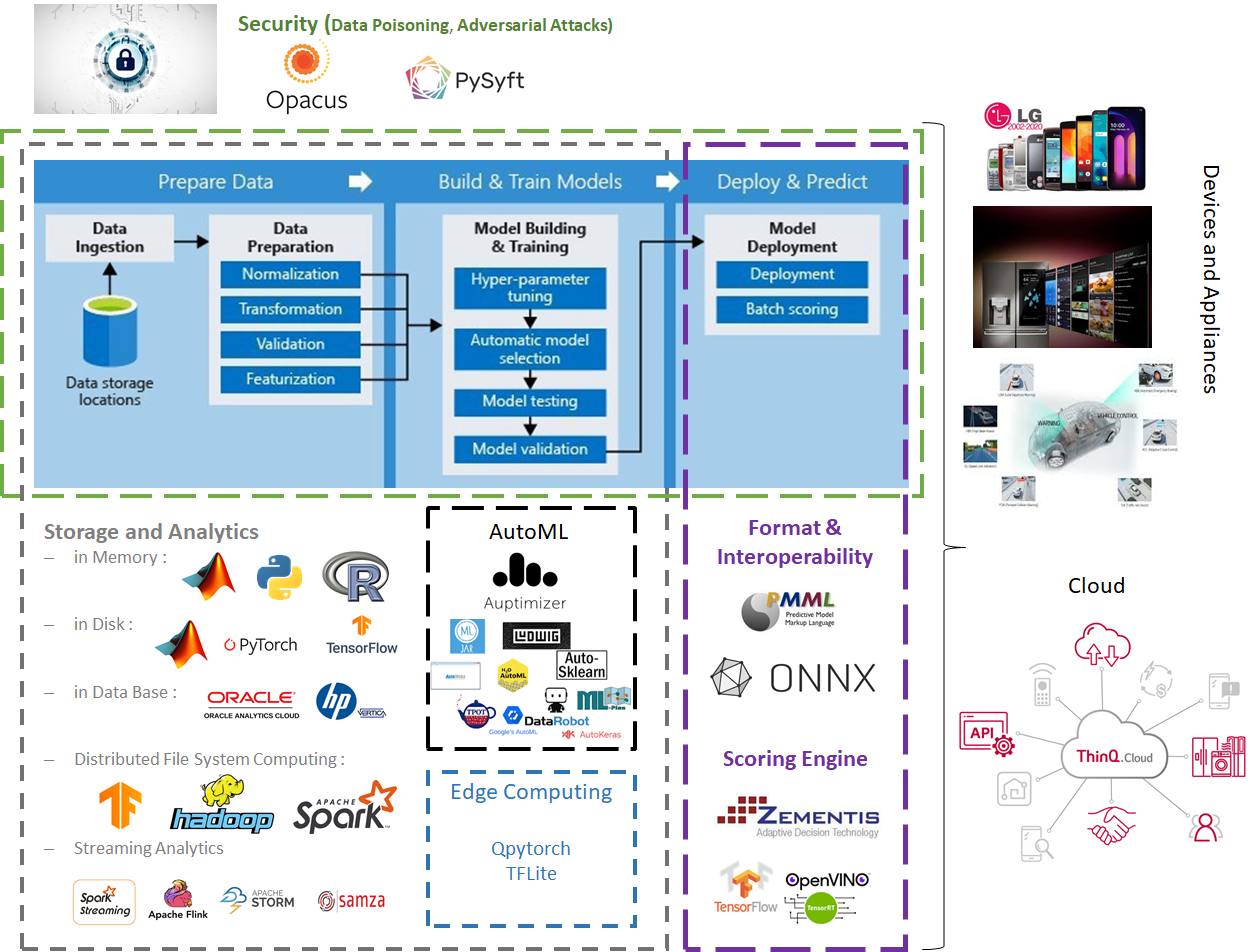}
\caption{AI Infrastructure typical in PdM Use-cases}\label{infra}
\end{figure}

\begin{figure}
\centering
\includegraphics[width=10.0cm]{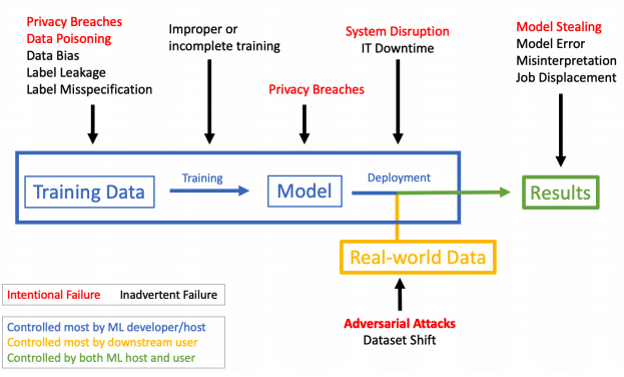}
\caption{Privacy Preserving Machine Learning. \scriptsize{Image source \url{https://pnptc-media.s3.amazonaws.com/images/machine-learning-security-1.width-800.png}}} \label{privacy}
\end{figure}

Most of the above discussion address the recent algorithmic developments for handling Predictive Maintenance Use-cases. In this section, we discuss the system/infrastructure-centric research pertaining to PdM use-cases. Basically there are several important aspects while dealing with a typical PdM use-case. These include,
\begin{itemize}
    \item[--] \textbf{Storage and Analytics} : This is an important aspect while architecting the PdM Pipeline. One of the major point that dictates this decision is the size/volume of data to compute during data preprocessing and model training (see Fig.~\ref{infra}). Most of the existing paradigms can be broadly categorized into one of the following approaches (see Fig \ref{infra}),
    \begin{enumerate}
        \item in-Memory analytics : This is the most widely used paradigm for small/medium scale data. In this setting, the data is loaded in memory (typically CPU or GPU) and sequential (or parallel) compute happens within a single node (computer). Typical examples include, Matlab, Python, R-based scripts. Recently, significant speed-ups in mathematical computation has been achieved through use of GPUs. Some typical tools that adopt this approach include Tensorflow \cite{tensorflow}, Pytorch \cite{pytorch}, MXNet \cite{mxnet} etc. on Cuda.
        \item in-Disk analytics : This mechanism allows similar batch processing of the data (as above), with the advantage that the entire data resides on disk (i.e. is not loaded in memory). The data is sequentially loaded to memory in mini-batches and processed. Typical examples include MATLAB's (mem-map), Tensorflow \cite{tensorflow}, Pytorch \cite{pytorch} (through appropriate design of data-loaders). This mechanism allows to process large scale data (which does not fit in memory); but is typically slow (owing to frequent disk read-write operations).
        \item in-Database analytics : This mechanism host the data in structured (SQL) or unstructured (NoSQL) databases and provide analytics within the data-base. The data resides in the database and compute is taken to the data. Some typical solutions available are, Oracle's (in-database analytics) \cite{oracle}, HP Vertica \cite{vertica}, Google Big-Table \cite{bigtable} etc. This mechanism allows to handle large scale data. But the adopted mechanisms are mostly parallel computing as opposed to distributed. Further the analytics suites are limited to proprietary ones developed by the DB providers. 
        \item Distributed File System Computing: Probably one of the most popular approach for handling big-data storage and compute is the distributed computing (using Spark \cite{spark}, Hadoop ) with data stored in Hadoop Distributed File System (HDFS). This eco-system has been greatly researched and developed and is one of the more mature approaches to build ML models using big-data (like, ADMML \cite{admml}, MLLIB \cite{Mllib}, PhotonML \cite{photonML} etc). Recently there have been a few newer technologies like (DataBricks) Horovod Runner \cite{horovod}, Intel's BigDL  \cite{BIGDL} (for Deep Learning on Spark) etc., or Visualization on Spark (using Tableau \cite{tableau} etc.) \item Streaming Analytics: Another aspect typically seen in PdM use-case is the need for real-time streaming analytics. In this paradigm the data is processed as it's received. There are several mechanisms to do that like native, or consumer-subscriber based analytics. Some of the prominent tools with a mature eco-system built around them include, Spark Streaming , Apache Flink , Apache Storm \cite{streaming}, Samza \cite{samza} etc. 
    \end{enumerate}
    \item[--] \textbf{AutoML} (Automated Machine Learning): An important aspect of successfully building good predictive models for event/error diagnostics is careful tuning of the underlying model's hyperparameters. Lately there has been immense research on Automated Machine Learning (AutoML) for generic machine learning problems. A recent white paper by SensiML highlights the advantage of AutoML in IoT use-cases \cite{sensiml}. So, an important consideration while designing the PdM infrastructure is adapting platforms that support AutoML mechanisms. Some of the publicly available solutions include, TPOT \cite{tpot}, GCP , Auto sk-learn, Auto Gluon , H2o.ai \cite{automl,autoMLcomparisons} etc. Unfortunately, most of these above solutions are tied to a specific platform (Pytorch, sk-learn or Tensorflow etc). Recently, LG Electronics open-sourced their Auptimizer Library \cite{auptimizer}; which provide platform agnostic solutions for AutoML. This provides a huge advantage for use-cases not tied to a single eco-system.     
    \item[--] \textbf{Security} : Privacy Preserving machine learning is of utmost importance while designing a PdM Infrastructure. Typical to most PdM use-cases the sensor data reflects the usage of a device, appliance etc. Building infrastructures that ensure compute and data privacy advocates ethical machine learning \cite{ethical} . There are several aspects to secured ML at the different phases of typical ML pipeline (see Fig \ref{privacy}). Some existing solutions to handle a few of these aspects include, Opacus \cite{opacus} , PySyft \cite{pysyft} etc.       
    \item[--] \textbf{Deployment}: Finally, a very big decision while designing an AI infrastructure for PdM is the deployment solutions available. Broadly the deployment solutions target a cloud-based or edge-centric deployment (see Fig \ref{infra}). There are two important aspects to consider here, 
    \begin{enumerate}
        \item Deployment Format : This indicates the format that the machine learning (or deep-learning) model needs to be represented before porting it to an inference engine. Some popular formats include, PMML \cite{pmml} (traditional ML Models), or ONNX \cite{onnx} (modern deep-learning models).
        \item Scoring Engine : This is the inference engine that runs on the cloud or edge device. This engine reads - in the data and using the provided model (in the deployment format), provides the prediction scores. Some of the popular solutions include, Adapa (Zementis) \cite{adapa} (for PMML), TFlite \cite{tensorflow} (for .tflite or ONNX), OpenVino \cite{openvino} , TensorRT \cite{tensort} (ONNX) etc. 
    \end{enumerate}
    The adopted / designed  infrastructure should be compatible or easily extensible to handle the above aspects in a PdM pipeline.
\end{itemize}

\section{Case-Studies} \label{sec_case_studies}

We have discussed value of PdM for different stakeholders and state of the art PdM techniques. In this section we present the publicly available case-studies which illustrate the value of PdM for electromechnical appliances in home and commercial environments. The example case-studies include refrigerators,  HVACs, washers, driers,  water filters and air purifies. This list is only a suggestive set and is used to illustrate the utility of PdM in the context of electromechanical appliances. 

\subsection{Predictive maintenance with the Azure IoT dataset }

Here we present a brief summary of our predictive maintenance and root cause analysis work using machine learning in IoT applications. Machine learning and modern deep learning methods are powerful tools for analyzing normal and abnormal behaviour of the IoT devices. We have investigated several standard datasets including MIMII (a sound-based industrial machine fault detection dataset~\cite{purohit2019mimii}), AI4I~\cite{ai4i}, and a predictive maintenance (PdM) dataset from Azure~\cite{azure_dataset}. Among these, the Azure predictive maintenance dataset is the most interesting, as its multivariate training data includes healthy data that changes over time to develop a failure. It is therefore a dataset that is suitable for building models to predict a future fault before it happens. We will therefore focus on this dataset in this document. 

\subsubsection{Description of the Microsoft Azure dataset}

The Azure dataset is a simulated dataset containing one year of hourly data for 100 devices with vibrations, rotation speeds, voltage, and pressure data, as well as time series records of when various parts were replaced on the devices. Each device has four parts which frequently fail. The devices are regularly inspected by technicians. Sometimes the technicians identify developing problems and replace the parts before they fail, but other times the parts fail before a human can replace them. Most parts will fail and need to be replaced a couple of times per year. All of these replacements are recorded, along with the age and model type, and five different kinds of ``error" flags that are occasionally logged. These error flags serve as warnings, but do not necessarily mean a fault has occurred which requires the replacement of a part. A year of hourly time series data for four physical quantities is also recorded, including the voltage, rotation speed, vibrations, and pressure.  The task becomes to predict in advance when a part will fail to help the human technicians avoid missing real problems. We consider the goals of predicting a failure up to three or seven days in advance, so that the technicians can be giving enough warning. Figure~\ref{fig:azure_pie} shows that a little under 2\% of parts will fail in a given three day period. 

\begin{figure*}[!htb]
  \centering
  {\includegraphics[width=1\textwidth]{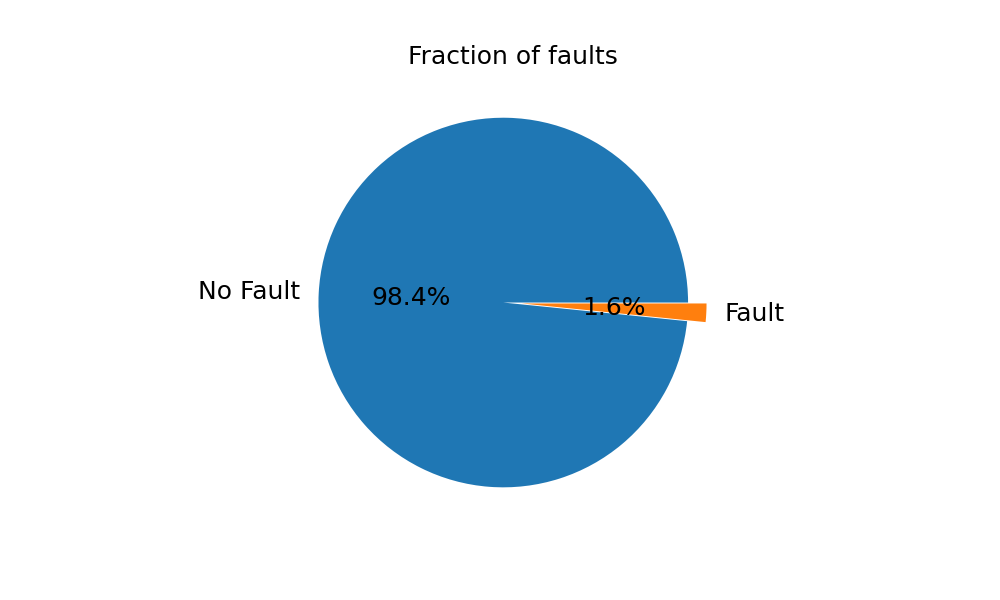}\label{fig:azure_pie}}
  \caption{Fraction of 3 day periods that contain of fault in the Azure dataset}
   \label{fig:azure_pie}%
\end{figure*}
\begin{figure*}[!htb]
  \centering
  {\includegraphics[width=1.0\textwidth, height = 0.4\textwidth]{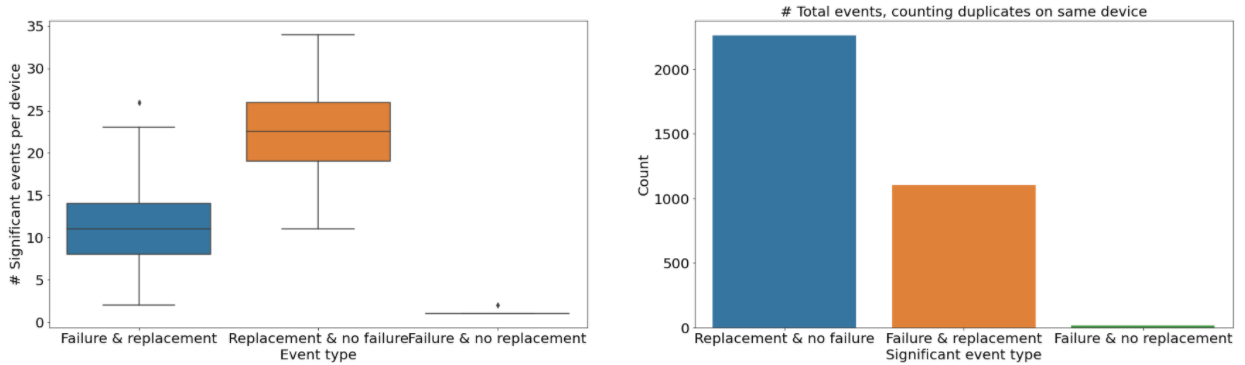}\label{fig:f13}}
  \caption{Illustration of the frequency with which humans successfully identify and replace before failures on a device (left) and the total number of replacements and failures across all devices (right).}
   \label{fig:azure_2}%
\end{figure*}

We also have information about the model type and age of each device. Figure~\ref{fig:azure_2} shows us that about two thirds of the parts are currently caught by humans before they fail, but and our algorithms will be attempting to catch many of the remaining third that they miss. 


\clearpage
\newpage
\newpage
\newpage

\begin{figure*}[!htb]
  \centering
  {\includegraphics[width=1.0\textwidth]{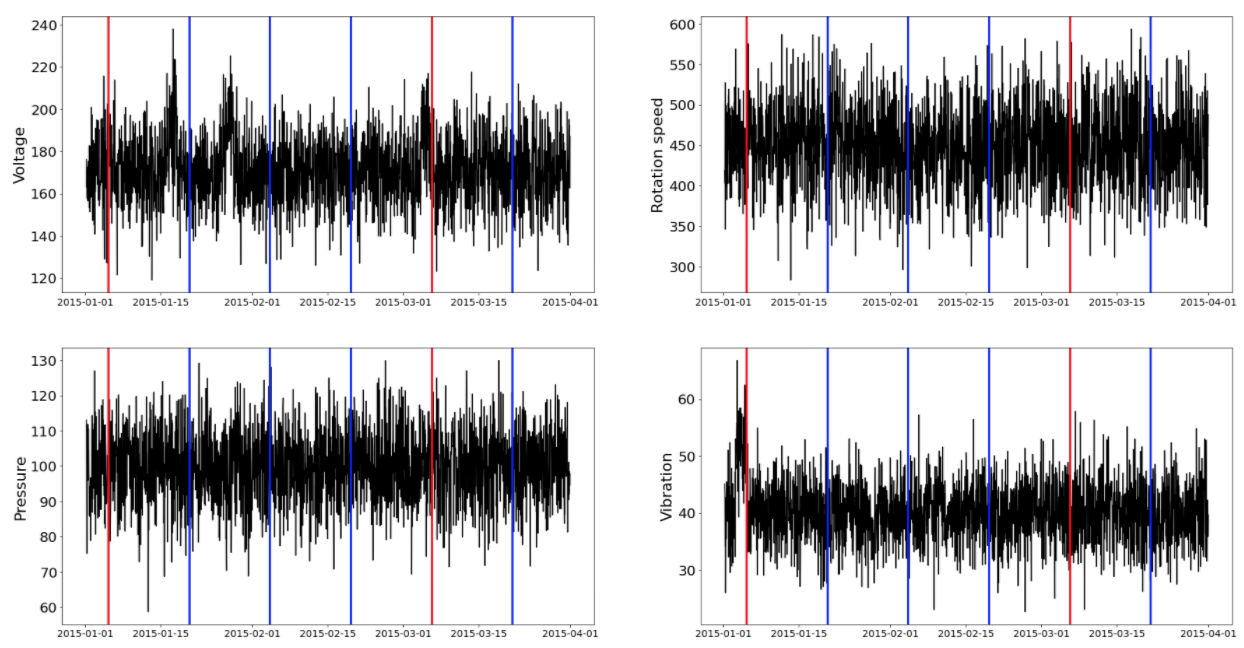}\label{fig:f19}}
  \caption{Metric values for device id=1. The red lines show part failures that were overlooked by humans, while the blue lines show parts that were replaced by humans before the failures occurred.}
   \label{fig:azure_8}%
\end{figure*}

\begin{figure*}[!htb]
  \centering
  {\includegraphics[width=1.0\textwidth]{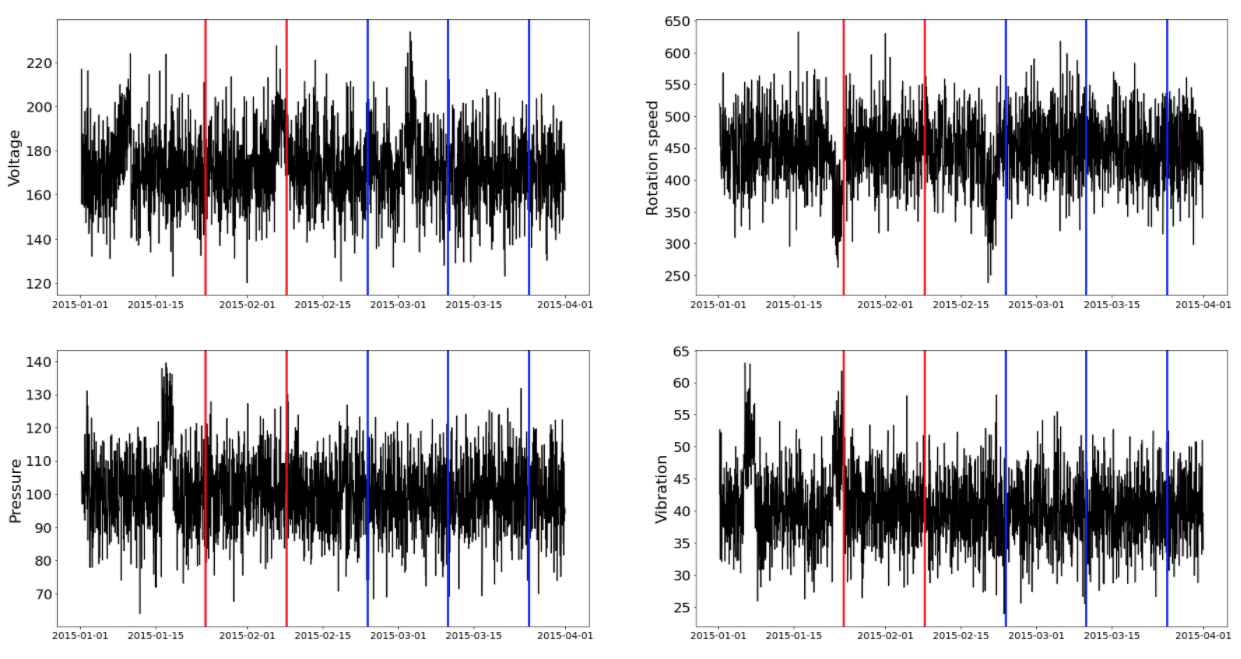}\label{fig:f20}}
  \caption{Same, but for a randomly selected different device (device 7). This one illustrates that not every fluctuation in a metric leads to a soon upcoming failure or a replacement. }
   \label{fig:azure_9}%
\end{figure*}


\begin{figure*}[!htb]
  \centering
  {\includegraphics[width=1.0\textwidth]{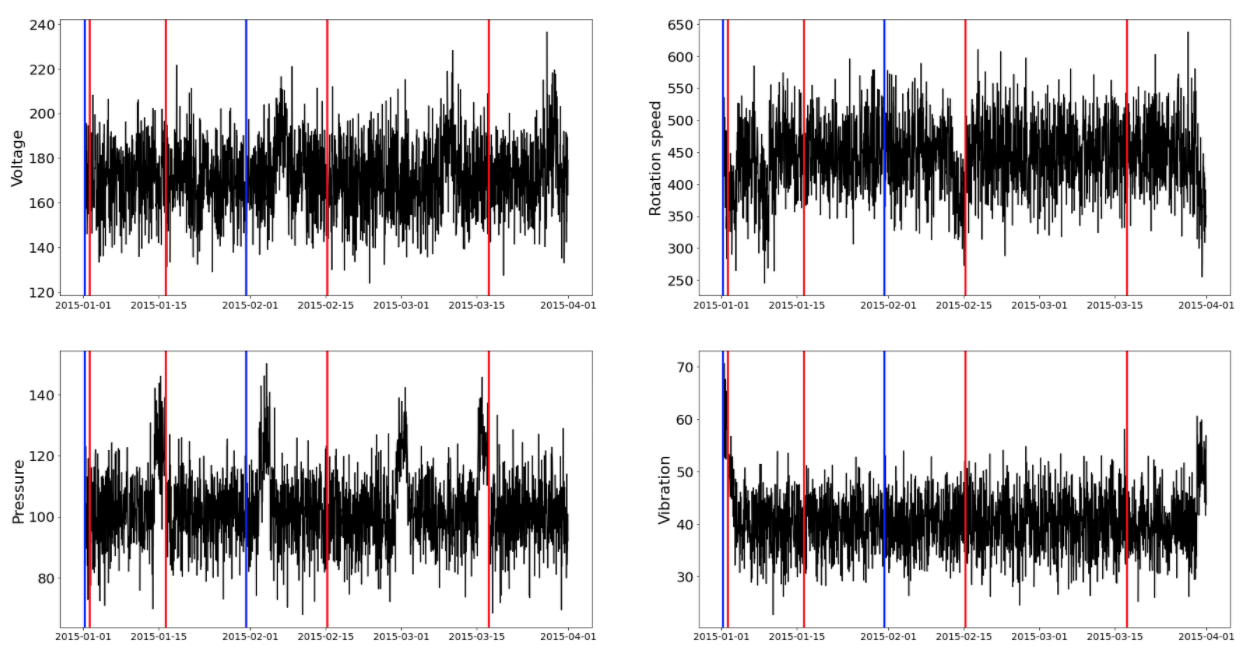}\label{fig:f21}}
  \caption{Same, but for another randomly selected device (device 17)}
   \label{fig:azure_10}%
\end{figure*}



\subsubsection{Data preprocessing and feature engineering}

The starting dataset is a natural one to use without too much preprocessing if the goal is to identify faults as they happen, but in our case the goal is to identify the faults before they happen (PdM). One of the most important prepossessing steps is therefore appending a new column to the dataset that records the number of days in the future until the part fails. Once this feature has been engineered it is then trivial to select custom cuts on it if our goal is to train a model to predict 1 day in the future, 3 days in the future, 7 days in the future, or whatever our business needs are. Similarly, it is reasonable to expect older parts are more likely to fail than newer parts, and that the time since the last part replacement will have predictive value for when the next replacement will be needed. We engineer this as an additional feature column as well. Figures \ref{fig:azure_8}, \ref{fig:azure_9}, and \ref{fig:azure_10} show the most important time series features for three example devices, along with vertical lines illustrating times when devices are replaced by humans before failure (blue) or when the device fails on its own before being caught (red). It is clear that anticipating a failure from the data can be challenging. Sometimes the failures are preceded by odd behaviors in one or more features, but not all do, and such strange behaviors do not guarantee upcoming failures. 

Standard machine learning feature normalization and cleaning of missing values is important of course, but for models which take time series as an input such as an LSTM or transformer model not much more feature engineering is needed than that. For traditional models which run on tabular data, however, it is necessary to do some further feature engineering to capture some key aspects of the time series nature. We therefore calculate new trailing aggregate features for each of the physical quantities (the voltage, rotation speed, pressure, and vibrations). These are the mean and standard deviation values of the features over the trailing month, as well as the ratio of those values in the most recent week to that month, and the most recent day to that month. Such features are very useful as encapsulations of what has changed recently which are important to predict an imminent failure, and are needed to assist a traditional model in doing time series predictions. But when using a model such as an LSTM it is best to let the model learn predictive time series structures on its own.

\subsubsection{Model predictions on the Microsoft Azure dataset}

We trained several classification to predict part failures three days in advance, ranging from traditional ML models such as SVMs and GBDTs, to deep learning models including LSTMs, GRUs, conditional variational autoencoders, and transformers. Perhaps surprisingly, the traditional models ended up performing much better than the deep learning models, illustrating that the most sophisticated models are not always the best, particularly when training sample sizes are not large.  Performances are shown in table~\ref{tab:AzurePerformances} for a GBDT model, both for the case of predicting failures three days in the future and for the case of predicting seven days in the future.

\begin{center}
\begin{table}
\centering
\begin{tabular}{ccccc} 
\hline
Number of days in the future & AUC & True positive rate & True negative rate & Precision \\ [0.5ex] 
\hline
3 & 0.987 & 0.911 & 0.965 & 0.300 \\ 
\hline
7 & 0.933 & 0.687 & 0.911 & 0.227 \\
\hline
\label{tab:AzurePerformances}
\end{tabular}

\caption{\label{tab:summary}Out of sample performances of the GBDT model when attempting to predict if a fault will happen three or seven days in the future.}
\end{table}


\end{center}

Two things immediately jump out from this table. First, the roughly 98\% AUC when attempting to predict faults three days in the future is impressive. But if business requirements only allow technician visits one day a week and predictions must be made seven days in the future the performance is considerably lower. Secondly, due to the extreme class (which is also typical of most fault detection problems), even a highly accurate model can be overwhelmed by false positives, leading to a very low PPV. To illustrate this, the false positive and false negative rates when predicting three days in the future over time are compared with the true positive rates in Figure~\ref{fig:azure_15}, under a scenario where the decision threshold is adjusted until the precision exceeds 50\%. We can see that even with a very accurate model such as this, a substantial fraction of the predicted faults will not actually fail during the next three days, and few true faults will be missed. But most actual faults are identified. To put this into greater context, however, we can compare all of this with the true negative rate, which we do in Figure~\ref{fig:azure_16}. Here we see that virtually all predictions say there will be no fault, and be correct about that. We have also made a public demo illustrating the results of the PdM model ~\cite{bib:AzureDemo}.

\begin{figure*}[!htb]
  \centering
  {\includegraphics[width=0.9\textwidth, height = 0.55\textwidth]{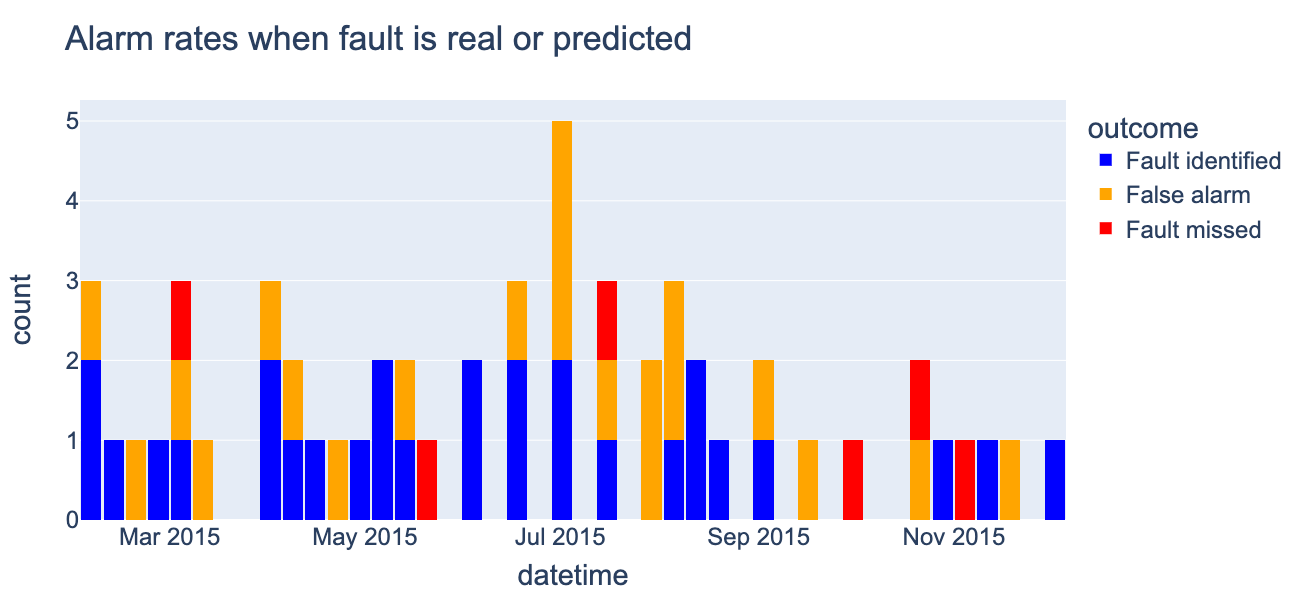}\label{fig:azure_15}}
  \caption{Alarm rates when fault is real or predicted}
   \label{fig:azure_15}%
\end{figure*}


\begin{figure*}[!htb]
  \centering
  {\includegraphics[width=0.9\textwidth, height = 0.5\textwidth]{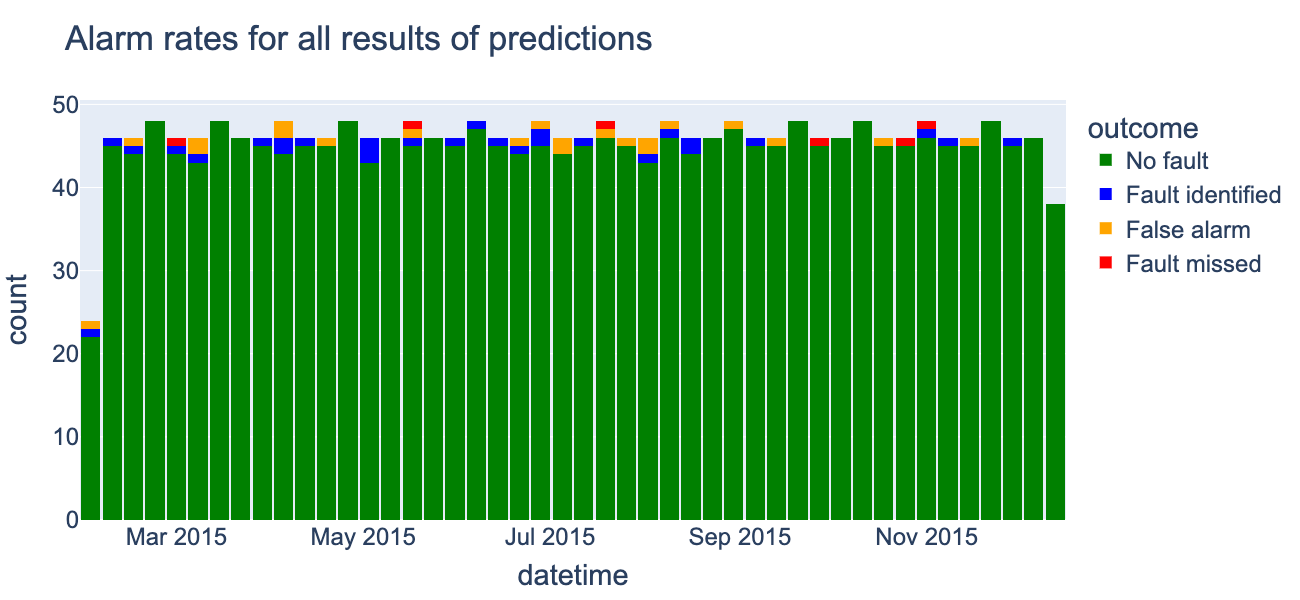}\label{fig:azure_16}}
  \caption{Alarm rates for all results of predictions}
   \label{fig:azure_16}%
\end{figure*}

\subsubsection{Causes of failures}

In addition, we want to do more than just predict when failures will happen. We want to be able to say why the algorithm thinks a failure is imminent. We developed a way to break down the amount that each feature considered by the algorithm contributes to the algorithm's decision. This method works by replacing the value of each feature by the median value across other example rows, one feature at a time, and recording the change in the predicted outcome to estimate the effects of that feature on the modeled results for that particular prediction.

Results of applying this methodology to a case where a fault was predicted to be imminent are shown in Figure~\ref{fig:azure_17}. The root cause analysis picks out the following top 10 features as driving the decision making:

The number of days since the last replacement of the fourth component of the device is shown to have a very large negative influence, pushing the model response more than 0.1 towards a fault determination by itself. The value of the average voltage over the last day and during the most recent hour both have almost as large of a negative influence, making a fault determination more likely. A variety of other features ranging from days since certain error flags were seen to rotation speeds play a small positive or negative role in the prediction, but the first three features carry by far the biggest importance in the decision making for this fault determination. Crucially, this determination comes out different for every fault. For a different example it might be the vibration speeds over the last week or the the device age that play the largest roles. 





\begin{figure*}[!htb]
  \centering
  {\includegraphics[width=0.8\textwidth, height = 0.6\textwidth]{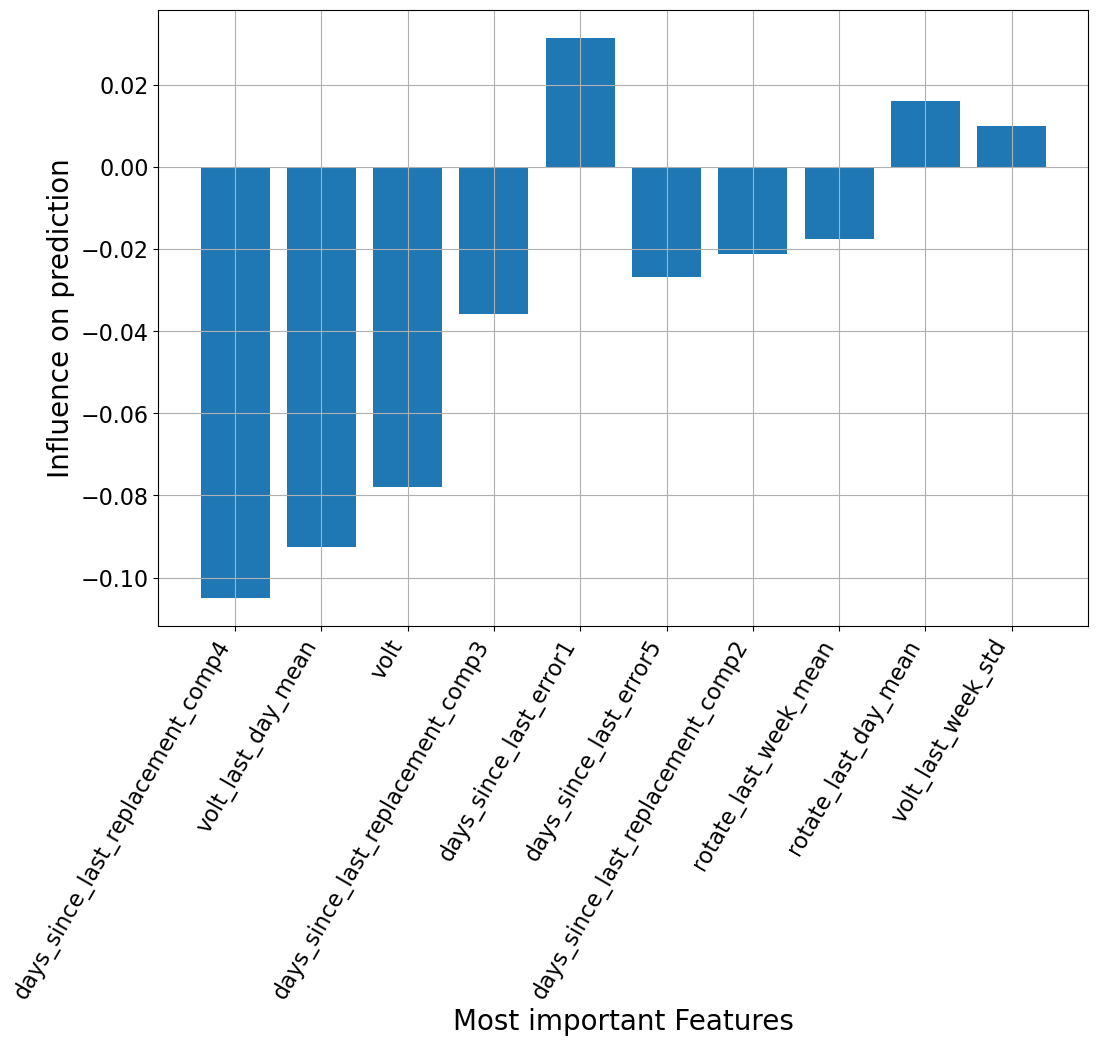}\label{fig:azure_17}}
  \caption{Root cause result for device1}
   \label{fig:azure_17}%
\end{figure*}
\subsection{Backblaze dataset}
We have conducted an experimental study where we used the services and tools offered by the Google Cloud Platform (GCP) to develop an end-to-end machine learning pipeline for anomaly detection problem in the context of hard disk failures. Through this study we evaluated the capabilities of GCP in managing large-scale machine learning problems specifically for the problem of anomaly detection using large-scale dataset. We have elaborately discussed about GCP and the services offered by it in subsection \ref{platforms}. For our study, we have used the Backblaze dataset \cite{backblaze}, discussed in subsection \ref{datasets} as it is one of the largest open-source datasets in this domain. On a high-level our machine learning pipeline consists of the following broad modules- data preprocessing and scrubbing, data ETL, model training and model evaluation. In this subsection we discuss in details the different modules of our end-to-end machine learning pipeline developed on the GCP platform. We also report the results and the findings of our experimental study.

\subsubsection{Data Preprocessing, Data Scrubbing and ETL}
We have used the Dataproc service offered by GCP to preprocess the Backblaze data so that it is amenable for the model development for the detecting anomaly in the hard disk data. As the Backblaze dataset is a large-scale dataset consisting of $228$ million records, we have used the PySpark distributed computing framework which is supported by the Dataproc service of GCP. Using PySpark and Dataproc, we have performed data cleaning operations like removing rows having missing data, removing features having more than $25\%$ of missing column data followed by normalization of values for all features having numeric values. Although Backblaze dataset is humongous, it is important to note that it is severely imbalanced with only $9000$ records out of $228$ million records having positive labels. Such extreme imbalance in input data can affect model performance. To overcome this problem, we use random downsampling to balance positive and negative samples of the data. Finally we upload our cleaned and preprocessed data to Google cloud storage (GCS) bucket so that the model development component of the pipeline can access it and subsequently train and evaluate the model.
\subsubsection{Model Development}
For the model development component of the pipeline, we have used the Vertex AI service of GCP. The model development component of the pipeline consists of two phases- model training and model evaluation. Vertex AI provides two methods for model development. The user can either use the AutoML service integrated in Vertex AI for the model development. AutoML provides an easy way for the users to develop the best model using model selection based on the input data and specified threshold. However, AutoML does not give the user to select any algorithm of their choice for the model development. For users requiring higher flexibility in terms of model selection, it allows users to develop their own custom training script and use it for the model development. We explored both the AutoML service as well as the custom training feature to evaluate the relative performance of the developed model. For custom training, we used three algorithms- random forest (RF), gradient boosted decision trees (GBDT) and multi-layer perceptron (MLP). We used Tensorflow $2$ for developing our custom training scripts. To expedite the training process on the large scale backblaze dataset, we have leveraged the distributed training feature offered by Tensorflow. For both the AutoML based training as well as out custom training, we have used a training and test split of $80\%:20\%$. Once the model is trained, we evaluate its performance on the test dataset. We present our findings in thee next subsection. 
\subsubsection{Results and Discussion}
\begin{table}[h!]
\centering
\begin{tabular}{ |c|c|c|c|c|c|c|c|c|c|c| } 
 \hline
 \textbf{Model} & \textbf{Accuracy} & \textbf{BCR} & \textbf{Precision} & \textbf{Recall} & \textbf{F1} & \textbf{ROC} & \textbf{PR} & \textbf{Model Cost} & \textbf{ETL Cost} & \textbf{Total Cost} \\ 
 \hline
 AutoML & $82.52$ & $78.24$ & $0.02$ & $73.97$ & $0.04$ & $0.86$ & $0.27$ & $\$148.07$ & $\$82.95$ & $\$231.02$ \\ 
 \hline
 RF & $86.98$ & $76.60$ & $0.02$ & $66.29$ & $0.05$ & $0.82$ & $0$ & $\$88.62$ & $\$82.95$ & $\$171.57$ \\ 
 \hline
 GBDT & $86.35$ & $76.61$ & $0.02$ & $66.93$ & $0.04$ & $0.84$ & $0.2$ & $\$88.56$ & $\$82.95$ & $\$171.51$ \\ 
 \hline
 MLP & $86.13$ & $73.36$ & $0.01$ & $61.8$ & $0.02$ & $0.18$ & $0$ & $\$88.42$ & $\$82.95$ & $\$171.37$ \\ 
 \hline
\label{tab:backblaze_performance}
\end{tabular}
\caption{Classification performance metrics and associated cost for different models developed in Vertex AI using Backblaze data. Classification metric values are represented in percentage and costs are represented in US dollars.}
\end{table}
To evaluate the performance of the developed models, we assessed their performance in terms of classical classification metrics as anomaly detection is a classification problem where the model classifies a sample as faulty and normal based on the feature values present in the data. We reported the measures for seven classification metrics- accuracy, balanced classification rate (BCR), precision, recall, F1 score, area under the curve for receiver operating characteristics (ROC AUC), area under the curve for precision and recall (PR AUC). In addition to the classification metrics we also reported three measures for cost analysis- model cost, ETL cost and total cost. Model cost refers to the cost incurred for training the model, ETL cost refers to the cost incurred for data extraction, transformation and loading and total cost refers to the sum of the costs incurred for model training and ETL. The first row in Table \ref{tab:backblaze_performance} reports the performance metrics and cost measures for the model developed using the AutoML service of Vertex AI and the next three rows in the table report to the performance metrics and cost measures for the model developed using custom training based on random forest, gradient boosted decision trees and multi-layer perceptron. In our experiment, the model selection algorithm of AutoML selected GBDT as the best performing model. Our held-out test data is extremely unbalanced as we only balanced the training data through random undersampling. Hence, the most important metrics for us in the context of anomaly detection for unbalanced test set is balanced classification rate (BCR) followed by recall, precision, F1 score, ROC AUC, PR AUC and finally accuracy. From Table \ref{tab:backblaze_performance}, it can be observed based on the top ranking classification metrics especially BCR and recall, that the AutoML model outperforms all the custom training models. The closest competitor to the AutoML model is the GBDT custom training model which reports a $2\%$ lower BCR and $7\%$ lower recall compared to the AutoML model. When we analyze the cost, however, we observe that AutoML model incurs a higher cost for model training. The ETL cost is the same for all  the  five models as we use the same data preprocessing component of the pipeline for all the models. The higher training cost associated with the AutoML training can be associated with higher number of  resources used by it during the training phase specifically during model selection and hyperparameter tuning. However, we can conclude from our results that AutoML model performs better in anomaly detection at a slightly higher cost compared to our best performing custom training model, GBDT.

\subsection{Failure Prediction on the Milling Dataset}

In this section, we provide different variations of variational autoencoder (VAE), as a Bayesian modeling which captures the uncertainty in the decisions, for failure prediction in the UC Berkeley milling data set~\cite{milling}. The milling dataset is includes $16$ cases of milling tools making cuts in metal. For data generation, six different combination of cutting parameters were considered: the metal type (either cast iron or steel), the depth of cut (either $0.75$ mm or $1.5$ mm), and the feed rate (either $0.25$ mm/rev or $0.5$ mm/rev). At the beginning of each case, a fixed set of the cutting parameters are selected to run the machine and collect the data. For instance, case one has a depth of cut of $1.5$ mm, a feed rate of $0.5$ mm/rev, and is performed on cast iron. Each case contains multiple individual cuts from when the tool is new (healthy) to worn (either degraded or failed). There are $167$ cuts amongst all $16$ cases. 

\subsubsection{Data Description}
\begin{figure}
    \centering
    \includegraphics{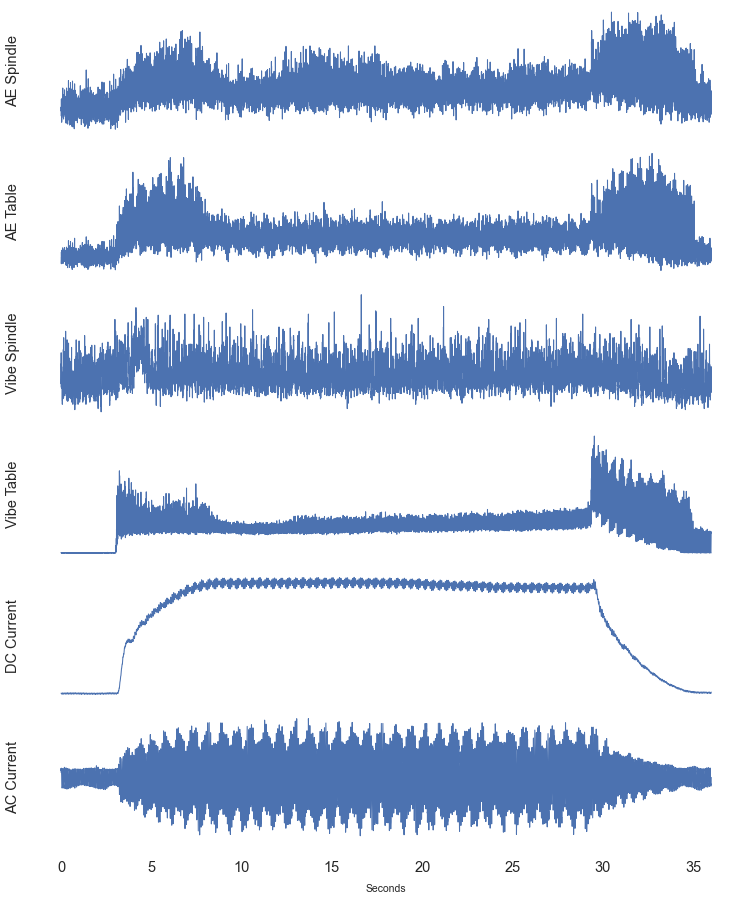}
    \caption{A sample of a single cut (cut 99) from Milling dataset}
    \label{fig:mill_example}
\end{figure}

Data is collected during each cut from the time the machine starts to the time it goes off. Throughout the cur, $6$ features are collected: acoustic emission (AE) signals from the spindle and table; vibration from the spindle and table; and AC/DC current from the spindle motor. Frequency of data collection is $250$ Hz and each cut has $9000$ data points. Figure~\ref{fig:mill_example} shows a representative sample of a single cut (cut 99). Each cut has a region of stable cutting, which is when the tool reaches its desired speed and feed rate, and fully engaged in cutting the metal. For the cut in Figure~\ref{fig:mill_example}, the stable cutting region is between sample point $2500$ and $7500$ when the tool leaves the metal it is cutting.

\subsubsection{Data Preprocessing}

For the Milling dataset, the pre-processing pipeline developed in~\cite{hahn2021self} is used. First, each sub-cut is normalized between zero and one. To make it amenable for disentangled variational autoencoder (VAE), all the sub-cuts in the input data should be of the same length. Hence, each sub-cut was zero-padded and centred to a final length of $2500$ samples. The final training, validation, or testing array, $X$, would have the following shape: $X = [$number of sub-cuts, length, number of features$]$, where the length is equal to the padded size of the sub-cuts ($2500$ samples) and the number of features is equal to one (the current signal). Each sub-cut was labelled either healthy (0), degraded (1) or failed (2) according to Table~\ref{tab:mill}. When a tool state is considered to be `wear', the prior 15 cuts were labelled as failed. The next 30 cuts were labelled as degraded. Cuts with tool breakage were removed from the data set. 

\begin{table}[h]
    \centering
    \caption{Labeling rule for each cut.}
    \begin{tabular}{lcc}
        \hline
        State & Label & Flank Wear ($mm$)  \\
        \hline
        Healthy & 0 & $0\sim 0.2$ \\
        Degraded & 1 & $0.2\sim 0.7$ \\
        Failed & 2 & $>0.7$ \\
        \hline
    \end{tabular}
    \label{tab:mill}
\end{table}

\subsubsection{Model}

\begin{figure}
    \centering
    \includegraphics[width=\textwidth]{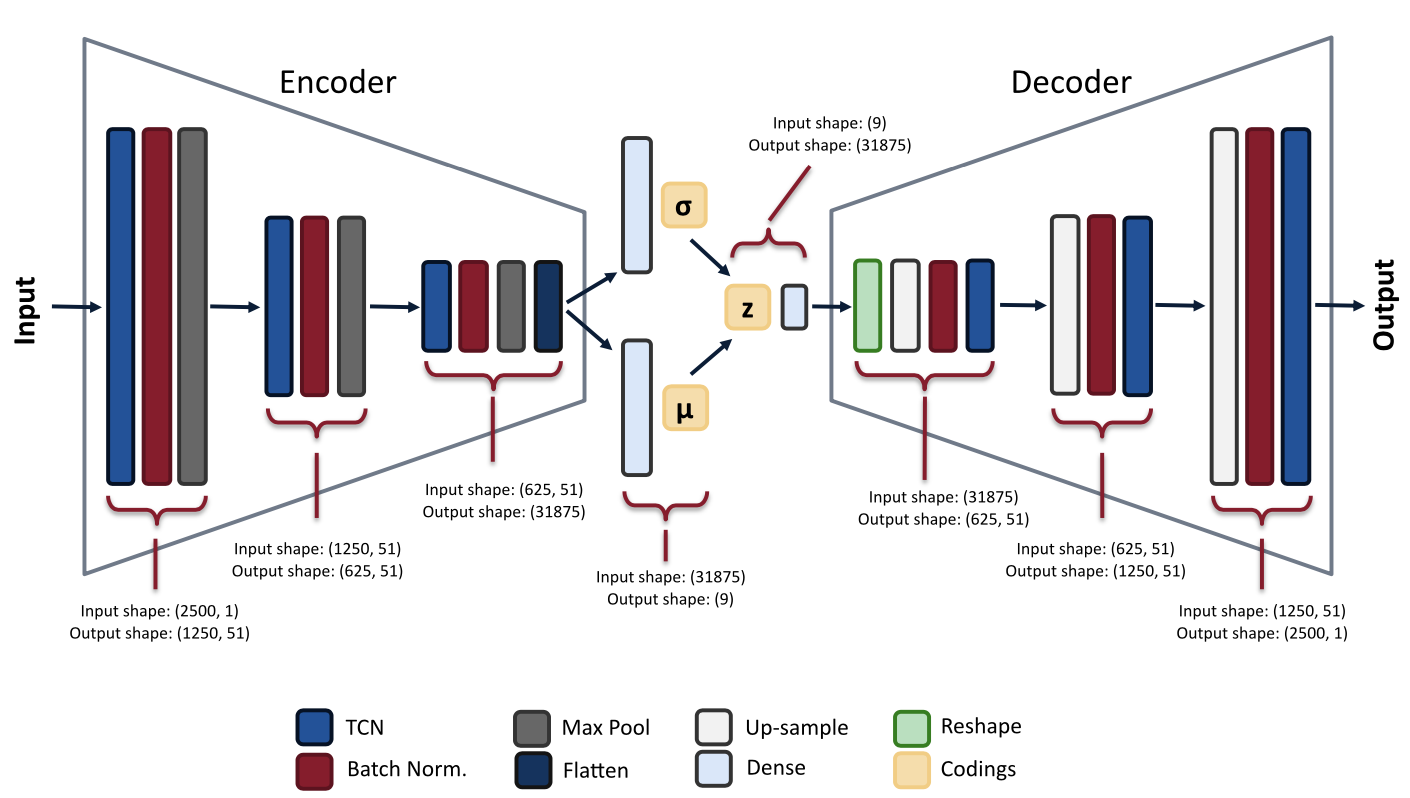}
    \caption{The VAE model architecture used for failure prediction~\cite{hahn2021self}.}
    \label{fig:vae}
\end{figure}

We use the same $\beta$-VAE model with similar architecture as in~\cite{hahn2021self}. The architecture, shown in~\ref{fig:vae}, includes three blocks of ``TCN block + batch-normalization + $2\times 2$ max-pooling layer", with activation functions being scaled exponential linear units (SELU). The output of the encoder is sent through a fully connected layer to generate the mean and standard deviation of the normal distribution of the latent space. The mean and standard deviations are used to sample from a Gaussian distribution to produce the coding $z$ which is processed through another fully connected network, called decoder. The decoder largely reverses what the encoder does. The data to the decoder is first reshaped, followed by a 2x up-sampling, and repeated twice more. 

\subsubsection{Results and Discussion}

\begin{figure}
    \centering
    \includegraphics[width=\textwidth]{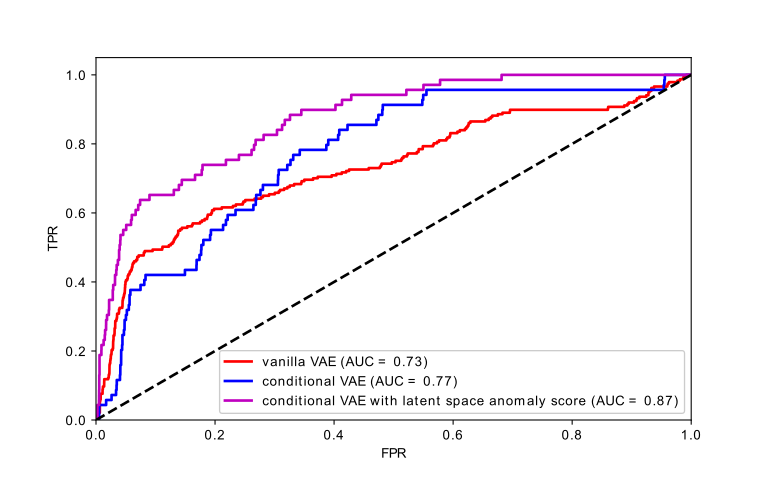}
    \caption{Comparison of VAE variant for failure prediction in Milling dataset.}
    \label{fig:vae_comp}
\end{figure}

One of the main advantages of using a VAE-based anomaly detection method is the Bayesian formulation of VAE. It allows the users to quantify the amount of uncertainty in their prediction of failure~\cite{wang2020advae,pol2019anomaly,an2015variational}. In vanilla VAE, the assumption is that the covariance matrix of the reconstruction output is fixed, e.g. identity matrix. This assumption limits the power that VAE can offer for anomaly detection in order to simplify the modeling and training of the VAE. One level of improvement on vanilla VAE is to predict the uncertainty in the output of VAE as well, which we call conditional VAE~\cite{wang2020advae}. This allows the user to have more clarity about the confidence of their decision, hence improving the performance. Next improvement is the idea of quantifying the anomaly score in the latent space of VAE~\cite{hahn2021self}. Since the latent space provides  disentanglement of parameters of the input space, it's been shown that it provides a better anomaly score which leads to higher AUC. Figure~\ref{fig:vae_comp} demonstrates a comparison of these three variants of VAE applied on the Milling dataset.

\section{External Resources}
\label{external_sources}

\subsection{Model Development Platforms}
\label{platforms}
Many real-world anomaly detection/ prediction problem involves humongous amounts of historical data. In such scenarios, it is important to have an end-to-end machine learning (ML) system which can scale well in presence of large-scale historical datasets and classify or predict anomalies in a time-efficient manner amenable for a real-world system. In this context, multiple cloud service providers like AWS, Google Cloud Platform (GCP) and Microsoft Azure offer different services which cater to the needs of users working with large-scale machine learning problems. In our study, we have explored each of these cloud platforms and investigated the services that they offer for developing an end-to-end machine learning pipeline for large-scale anomaly detection problem. In this section we will discuss about each of these cloud service providers and the relevant services that they offer for building such large-scale machine learning systems. 
\subsubsection{Google Cloud Platform}
GCP offers multiple services for large-scale machine learning problems. Some of the most important services in this regard are Dataproc, AutoML tables and Vertex AI. 

\textbf{Dataproc}\\ 
Dataproc is a managed Spark and Hadoop service that enables users to leverage the benefits of open-source data tools for batch processing, querying, streaming and machine learning. Dataproc automation allows fast cluster creation and management. It also offers economic advantage to users by allowing them to turn off clusters when they are not. actively used. Dataproc offers the following salient features which makes it a suitable fit for the data preprocessing module of our pipeline.

\textit{Low Cost.} This service is quite economic as it is priced at only $1$ cent per virtual CPU per hour in addition to the other cloud platform resources that can be utilized. Dataproc clusters also include preemptible instances that have lower compute prices, reducing user cost to a greater extent. Another advantage offered by the Dataproc service is that it only charges the users for what they have actually used using a second-by-second billing and a low, one-minute-minimum billing period instead of rounding up the charge to the next hour.

\textit{Super fast.} Dataproc significantly expedites the process of Spark and Hadoop cluster creation in comparison to creating them on-premises or through IaaS providers. In general, Dataproc clusters take much less time to start, scale, and shutdown thus accelerating the entire preprocessing pipeline. This enables the users to spend less time waiting for clusters and invest more time working with the data.

\textit{Integrated.} Dataproc enables users to seamlessly use other Google Cloud Platform services, such as BigQuery, Cloud Storage, Cloud Bigtable, Cloud Logging, and Cloud Monitoring through it's built-in integration with these services. Consequently, the user has a complete data platform at his disposal instead of just having a Spark or Hadoop cluster. Dataproc, for instance, can be effortlessly used for extraction, transformation and loading large amounts of data directly into BigQuery for the purpose of reporting.

\textit{Managed.} Dataproc enables the users to use Spark and Hadoop clusters without the assistance of an administrator or special software. The user can easily interact with clusters and Spark or Hadoop jobs through the Google Cloud Console, the Cloud SDK, or the Dataproc REST API. When the user has completed all his tasks on a cluster, the user can simply turn the cluster off so that he is not charged for an idle cluster. Additionally, the user does not have to worry about losing data as Dataproc is integrated with Cloud Storage, BigQuery, and Cloud Bigtable.

\textit{Simple and familiar.} The user is not required to learn new tools or APIs to use Dataproc thus facilitating an easy transition of existing projects into Dataproc without redevelopment. With frequent updates of Spark, Hadoop, Pig, and Hive, the user can be productive faster.

\begin{figure}
\centering
\includegraphics[width=15.0cm]{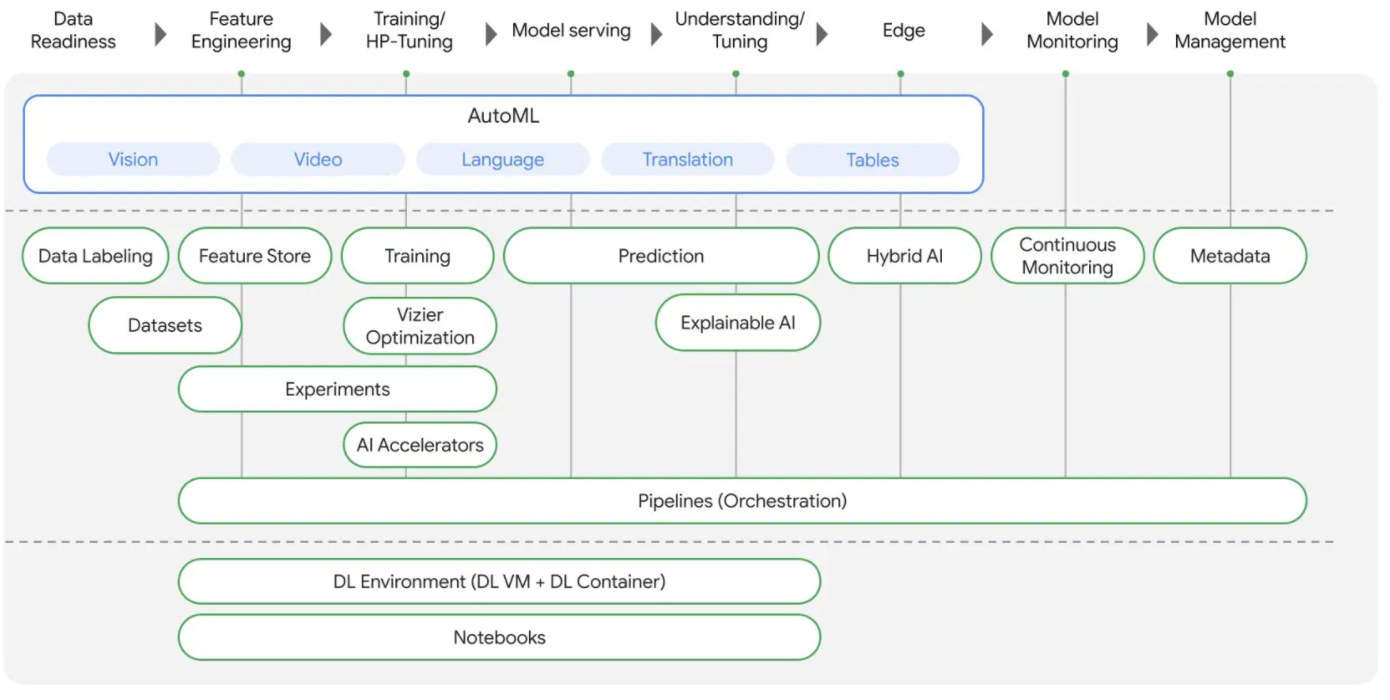}
\caption{System design of Vertex AI (Retrieved from https://techcrunch.com/2021/05/18/google-cloud-launches-vertex-a-new-managed-machine-learning-platform/).} 
\label{vertex_ai_system_design}
\end{figure}

\textbf{Vertex AI}\\
Vertex AI unifies two services- AutoML and AI Platform and creates an integrated API, client library and user interface. Vertex AI enables the user to utilize the AutoML training as well as a fully custom training according to his requirements. Irrespective of the option chosen by the user, Vertex AI allows the user to save models, deploy models and perform batch predictions on the test data.
Vertex AI supports different phases of machine learning pipeline development including creation of dataset and uploading data, model training, model evaluation and hyperparameter tuning, storing model and its deployment to an endpoint for production serving, sending prediction request and managing models and end points. Figure \ref{vertex_ai_system_design} illustrates the high level system design for Vertex AI. Vertex AI supports the following salient features which we summarize below.

Vertex AI supports AutoML using which the user can develop high quality custom machine learning moddels which can leverage Google's state-of-the-art transfer learning and hyperparameter search without having to actively develop high quality custom training routines. Vertex AI allows users to instantiate a virtual machine image containing the most popular AI frameworks on a Compute Engine instance without being concerned about the software compatibility. Vertex AI supports the creation, management and connection to virtual machines with JupyterLab which is the standard workbench for data scientists. Virtual machines come pre-installed with deep learning frameworks and libraries.
It also offers Vertex Matching Engine which offers a massively scalable, low latency, and cost-efficient vector similarity matching service. Vertex AI offers a data labeling service to get highly accurate labels from human labelers for better machine learning models. Deep Learning Containers in Vertex AI enables users to build and deploy models in a portable and consistent environment for all their AI applications in a time-efficient manner.
Edge Manager in Vertex AI seamlessly deploys and monitors edge inferences and automated processes with flexible APIs. Vertex AI promotes explainability by creating a deep understanding and trust in the user's model predictions with robust, actionable explanations integrated into Vertex Prediction, AutoML Tables, and Notebooks. Vertex AI hosts a feature store which contains a fully managed rich feature repository for serving, sharing, and reusing ML features. Vertex AI also supports artifact, lineage, and execution tracking for machine learning workflows, with an easy-to-use Python SDK. Vertex AI features automated alerts for data drift, concept drift, or other model performance incidents which may require user. supervision. Vertex AI supports neural architecture search by building new model architectures targeting application-specific needs and optimizing the user's existing model architectures for latency, memory, and power with this automated service leveraging Google’s leading AI research. Vertex AI enables users to build pipelines using TensorFlow Extended and Kubeflow Pipelines, and leverage Google Cloud’s managed services to execute scalability and pay per use. Users have the option to streamline their MLOps with detailed metadata tracking, continuous modeling, and triggered model retraining. Vertex AI enables users to deploy models into production in a seamless manner with online serving via HTTP or batch prediction for bulk scoring. Vertex Prediction offers a unified framework to deploy custom models which can be trained in a variety of frameworks like TensorFlow, Scikit or XGB, as well as BQML and AutoML models as well as on a broad range of machine types and GPUs. Vertex AI has a visualization and tracking feature called Vertex Tensorboard which can be immensely helpful for ML experimentation for visualizing model graphs in the form of images, text, and audio data. Vertex AI offers a set of pre-built algorithms as well as allows users to deploy their custom code to train models. Hence, Vertex AI offers a fully managed training service for users who need higher flexibility and customization or for users running training on-premises or through another cloud environment. Vertex AI also features Google Vizier for users to optimize the hyperparameters for their training model which can lead to maximum predictive accuracy.

\textbf{AutoML Tables}\\
AutoML tables is a supervised learning service which allows the user to train a machine learning model with example labeled data. AutoML Tables utilize tabular (structured) data to train a machine learning model which makes predictions on previously unseen data. The model learns to predict the target which is one specific column from the dataset. The model will learn the underlying function from a subset of the other data columns which are considered as input features. The user has the flexibility of building multiple kinds of models using the same input features just by altering the target. AutoML table follows a specific workflow which starts by gathering the data, data preparation, training, evaluation, testing, deployment and prediction. AutoML tables are capable of creating thee necessary model for solving binary classification, multi-class classification or regression problem. AutoML table automatically determines the problem and the relevant model that has to be created based on the datatype of the target column. The main features of the AutoML tables are summarized as below:

\textit{Data Support.} AutoML Tables enable users to create clean and effective training data by offering useful insights about missing data, correlation, cardinality, and distribution for each of the existing features in the dataset. AutoML tables do not charge the user for the data import and analytics service. The user is only charged when he commences the model training.

\textit{Feature Engineering.} Once the model training is triggered by the user, the AutoML tables automatically perform common feature engineering tasks which include normalization and bucketing of numerical features, create one-hot encoding and embeddings for categorical features, performing basic data processing for text features and extraction of date and time related features from timestamp columns. 

\textit{Model Training.} AutoML tables support parallel model testing. When model training is initiated by the user, AutoML tables takes the dataset as input and commences parallel training using multiple model architectures. The feature allows AutoML tables to identify the best model architecture for the dataset under consideration without having to perform these tests sequentially over many model architectures. AutoML tables at present test over five model architectures- linear, feedforward deep neural network, gradient boosted decision trees, AdaNet and ensembles  of various model architectures. AutoML table uses the training and validations sets provided by the user to determine the best model architecture for the dataset under consideration. Subsequently, AutoML trains two models using the parameters and architecture which were determined during the parallel testing phase- 1. A model is trained using the training and validation datasets  provided by the user. The test set is then used to evaluate the developed model on the test set and 2. A model which is trained with the training, validation and test sets which is then used to offer predictions on new data. 

\textit{Model Transparency and Cloud Logging.} The user has the ability to visualize the structure of the AutoML tables model using the cloud logging service. Usinig the logging service, the user can visualize the hyperparameters as well as the hyperparameters and objective values which are used during model validation.

\textit{Explainability.} AutoML table promotes explainability by offering two primary ways to the user for gaining insight about the model and its operations- feature importance and test data export. The user can export the test data as well as the predictions made by the model. This capability enables the user to have insight into how the model is performing on each individual row of the training data. The user can examine the test set and the results through this capabilty which can help the user understand what types of predictions the model performs poorly so that the data can be improved to develop a higher quality model. 
\subsubsection{Amazon Web Services}
\begin{figure}
\centering
\includegraphics[width=15.0cm]{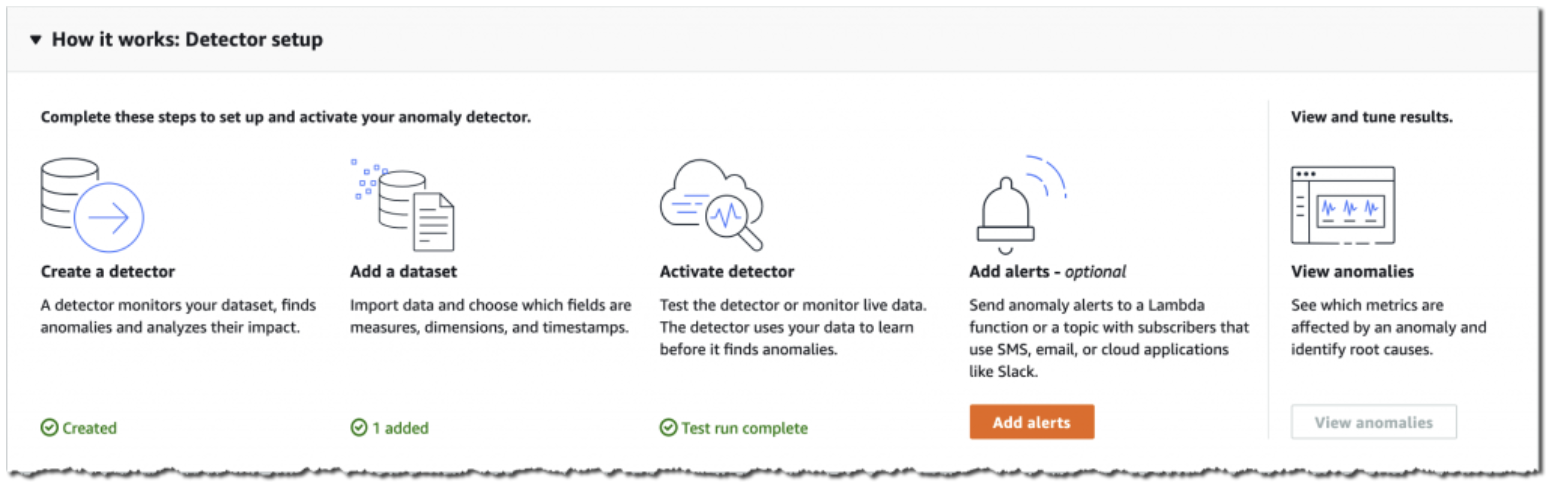}
\caption{High level system design for Amazon Lookout for Metrics (Retrieved from https://aws.amazon.com/blogs/aws/preview-amazon-lookout-for-metrics-anomaly-detection-service-monitoring-health-business/).} 
\label{aws_system_design}
\end{figure}

Amazon web services (AWS) has recently launched a new service called Amazon Lookout for Metrics that leverages machine learning to monitor the metrics which are important for business with greater speed and accuracy. This service also enables the users to gain insights into root cause analysis by diagnosing factors like unexpected dips in revenue, high rates of abandoned shopping carts, spikes in payment transaction failures, increases in new user sign-ups, and many more. Lookout for Metrics enables developers to set up autonomous monitoring for important metrics to detect anomalies and detect their root cause leveraging the same technology used by Amazon internally to detect anomalies in its metrics. The users or developers do not need any specific machine learning experience to use this service. Figure \ref{aws_system_design} illustrates the high level system design of Amazon Lookout for Metrics service.

Organizations in various industries aim to improve the efficiency in their business through technology and automation. While challenges can be different, identifying defects and opportunities early can often can result in material cost savings, higher margins, and overall better customer experience. Organizations have historically relied on manual audits of large amounts of data. However, such a solution is not scalable and is prone to human errors. Others use rule-based methods based on arbitrary ranges. These systems are static and are not adaptable to seasonality and other temporal changes, and can lead to numerous false positives. When anomalies are detected, developers, analysts, and business owners end up spending significant time attempting to detect the root cause of the change. ML can be an effective and transformational tool in these situations. However, applying ML algorithms to this problem requires careful model selection, training, testing, and deployment of the trained model for each type of data thus requiring a team of ML experts.

Amazon is a data-driven company, with a growing number of businesses that need to excel in terms of the health of their business, operations, and customer experience. The principal part of this effort has involved building and improving ML technology to detect anomalies in key performance indicators (KPI) such as website visits from different traffic channels, number of products added to the shopping cart, number of orders placed, revenue for every product category, and more. Amazon Lookout for Metrics gives access to the developers to the same ML technology used by Amazon. It detects anomalies in the user's data, clusters them intelligently and helps users to visualize aggregated results, and automates alerts. As it is a fully managed service, it manages the whole ML process so that the users can emphasize on their core business requirements. Additionally, the service improves model performance continually by incorporating real-time feedback on the accuracy and relevance of the anomalies and the root cause analysis.

Users can launch Lookout for Metrics very easily from the AWS management console. The users can connect their data to the service through the built-in data source integrations. As the next step, Lookout for Metrics trains a custom model for the user's data and finally it detects anomalies for the users to review and take corrective measures to handle them. Lookout for Metrics continuously monitors data stored in Amazon Simple Storage Service (Amazon S3), Amazon Relational Database Service (RDS), Amazon Redshift, Amazon CloudWatch, or SaaS integrations supported by Amazon AppFlow such as Salesforce, Marketo, Google Analytics, Slack, Zendesk, and many more. During this phase, the user is allowed to flag each field in the dataset as a measure (or KPI), dimension, or timestamp. Once the user's data source is configured and connected, Lookout for Metrics analyzes and prepares the data and performs algorithm selection to build the most accurate anomaly detection model. This detector runs on the user's data at a configurable speed and provides an adjustable threshold for sensitivity. 

When detecting an anomaly, Lookout for Metrics helps the user to emphasize on what matters the most by assigning a severity score for prioritization. It intelligently clusters anomalies that may be related to the same incident and summarizes the different sources of impact to enable the users to detect the root cause. Moreover, the user can configure an automatic action such as sending a notification via Amazon Simple Notification Service (SNS), Datadog, PagerDuty, Webhooks, or Slack or he can trigger a Lambda function to temporarily hide a product on their e-commerce site when a potential pricing error is detected. Domain knowledge and expertise are instrumental in determining if a sudden change in a metric is expected or is an anomaly. Lookout for Metrics enables users to provide real-time feedback on the relevance of the detected anomalies, promoting a powerful human-in-the-loop mechanism. This information is used as input to the anomaly detection model to improve its accuracy.

\subsubsection{Microsoft Azure}
Microsoft Azure has recently launched the Anomaly Detector service. Anomaly detector is an AI service that helps users foresee problems in their business or product before they occur. Anomaly detector can help users to enhance the reliability of their business by detecting problems early and adopting corrective measures to handle the problems beforehand. Using the Anomaly detector service enables the users to easily embed time-series anomaly detection capabilities inside user's apps to help them detect anomalies quickly. Anomaly Detector ingests time-series data of all types and selects the best anomaly detection algorithm for the user's data to ensure high accuracy. Anomaly detector is able to detect deviations from cyclic patterns and trend changes leveraging both both univariate and multivariate APIs. The service is flexible as it can be customized to detect any level of anomaly. The service can also be deployed either in the cloud or at the intelligent edge depending on the user's requirements. Next, we highlight the salient features of this service.

\begin{figure}
\centering
\includegraphics[width=15.0cm]{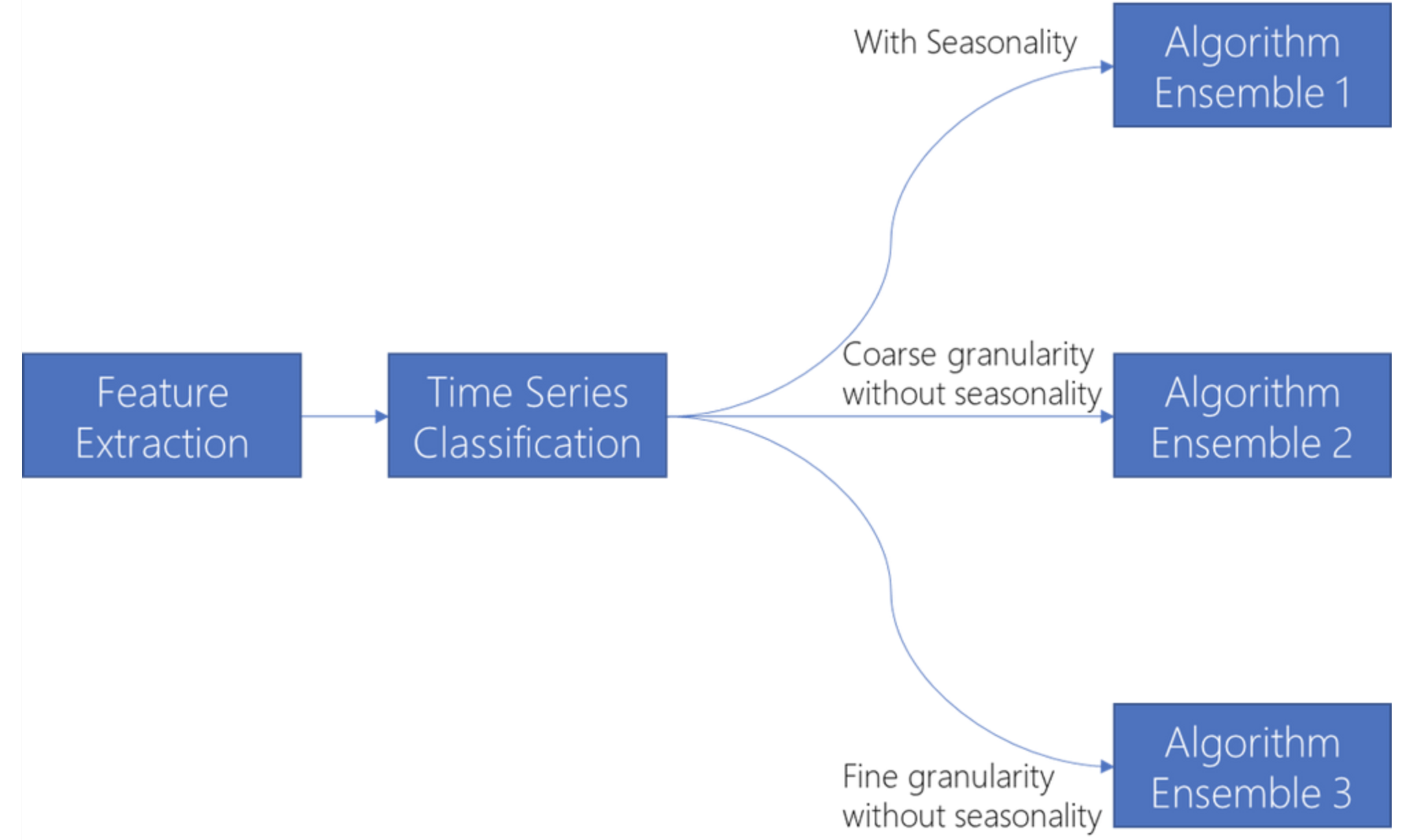}
\caption{Algorithm Selection flow of Anomaly Detector (Retrieved from https://techcommunity.microsoft.com/t5/ai-customer-engineering-team/introducing-azure-anomaly-detector-api/ba-p/490162).} 
\label{selection_framework}
\end{figure}

\textit{A gallery of algorithms.} State-of-the-art anomaly detection system often uses a universal approach which means they apply some specific algorithm on all types of time series. However, each algorithm can handle some specific type of time series better than the others. This service provides a generic framework to select different algorithm ensembles to handle a wide spectrum of different time series. The user has the option to choose from following algorithms- Fourier Transformation, Extreme Studentized Deviate (ESD), STL Decomposition, Dynamic Threshold, Z-score detector and SR-CNN. Figure \ref{selection_framework} illustrates the algorithm selection workflow of Anomaly Detector.

\textit{Detecting all kinds of anomalies through one single end-point.} In addition to common changes like spikes and dips, Anomaly Detector also detects other kinds of anomalies, such as trend change and off-cycle softness all of which are included in one single API endpoint.

\textit{Maintain simplicity outside- One parameter tuning.} Anomaly Detector service relies on one parameter strategy to make the tuning easier. The main parameter that the user needs to customize is “Sensitivity”, the value of which can vary from 1 to 99 to adjust the outcome according to the scenario. Other advanced parameters can be explored by data scientists for further tuning.

\begin{figure}
\centering
\includegraphics[width=15.0cm]{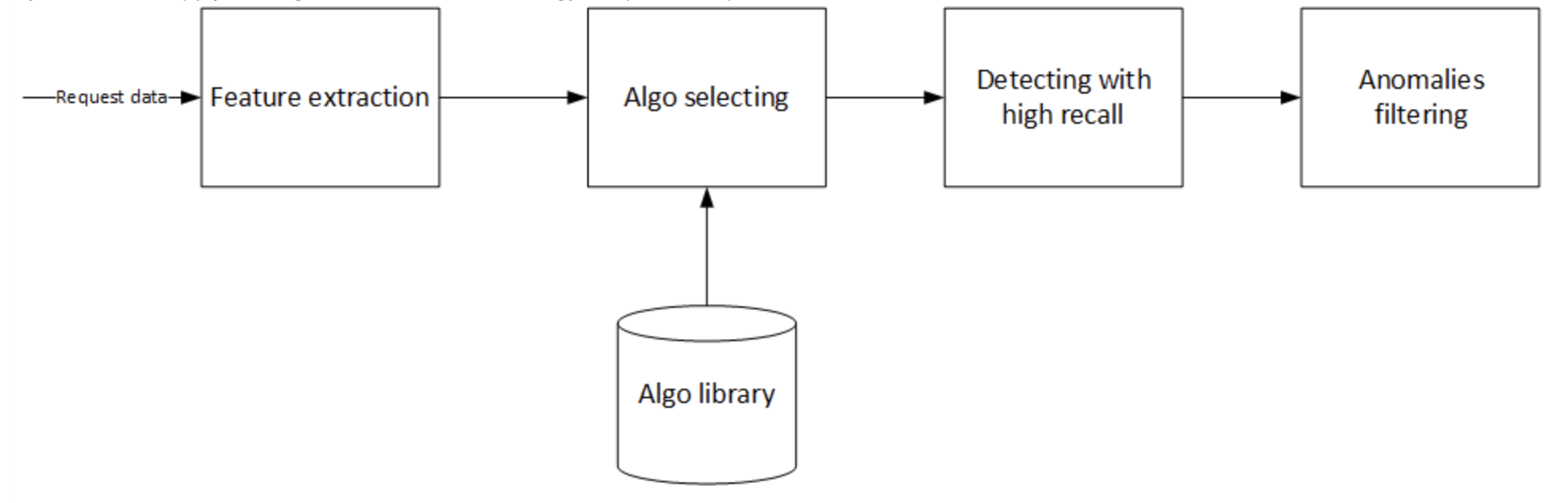}
\caption{High level System Design of Anomaly Detector (Retrieved from https://techcommunity.microsoft.com/t5/ai-customer-engineering-team/introducing-azure-anomaly-detector-api/ba-p/490162).} 
\label{system_design}
\end{figure}

\textit{Maintain sophistication inside- Selecting/ Deleting/ Filtering.} To support one parameter tuning, Anomaly Detector service assembled a gallery of anomaly detection algorithms and designed a sophisticated strategy to leverage the integrated power of machine learning for achieving good precision and recall. When the service receives the time series from an API request, it first extracts features from the time series, for example continuity, mean, STD, trend, and period. The system selects the most relevant algorithm with the help of those extracted features. It is difficult to strike a good balance in recall and precision in one single detection without enough context. Anomaly Detector segregates improving recall and precision into separate steps. Initially, it ensures that the recall is high. Then, it leverages those features, which are extracted from the time series combined with the sensitivity provided by the caller, to apply filtering on the results. This strategy has proved efficient and is able to offer better results. Figure \ref{system_design} illustrates a high level system design of the Anomaly Detector service.

\subsubsection{Other Service}
\cite{YangAMAZ21} proposes an anomaly detection algorithm selection service especially for IoT stream data based on Tsfresh tool and genetic algorithm. Anomaly detection algorithms (ADA) have been popularly used as services in many maintenance monitoring platforms. However, there exist various algorithms that could be applied to these dynamically changing data. Furthermore, we can commonly observe the phenomenon of conception drift in IoT stream data due to its dynamic nature. Therefore, it is challenging to select a suitable anomaly detection service (ADS) in real time. For accurate online anomalous data detection, this work developed a service selection method to select and configure ADS at run-time. This work uses two types of methods for feature selection- a time-series feature extractor (Tsfresh) and a genetic algorithm-based feature selection method. These methods are applied to rapidly extract dominant features. These features are very important as they act as representation for the stream data patterns. In addition to this, in this work, stream data and various efficient algorithms are accumulated as the historical data. This method uses a fast XGBoost based classification model which is trained to record stream data features to detect appropriate ADS dynamically at run-time. These methods play instrumental role in selecting suitable service and their respective configuration according to the patterns of stream data. The main success factor of this framework evolves from the features that aree used to depict and reflect the time-series data’s intrinsic characteristics. Consequently, experiments are conducted to evaluate the effectiveness of features selected by the genetic algorithm. Experimentations on both artificial and real datasets demonstrate that the accuracy of the proposed method surpasses various advanced approaches and can efficiently select the best service in different scenarios.

\subsection{Open Source Datasets for Predictive Maintenance} 
\label{datasets}
The UK-DALE dataset \cite{Kelly2015, UKERC2017} - domestic appliance-level electricity demand and whole-house demand from five UK homes. This is an open source data of energy usage in home environments with appliance-by-appliance consumption information. Energy desegregation is a computational technique for estimating appliance-by-appliance energy consumption from a whole-house meter signal. Training  desegregation algorithms using supervised learning approaches, researchers require labelled datasets
with energy usage from individual appliances. The UK-DALE dataset, an open-access dataset from the UK recording Domestic Appliance-Level Electricity at a sample rate of 16kHz for the whole-house and at 1/6 Hz for individual appliances. This data was recorded from five houses, one of which was recorded for 655 days. 

\cite{2020arXiv200313213L} lists several public data sets. These datasets could be leveraged for develping PdM models.

A new dataset of industrial machine sounds for malfunctioning industrial machine investigation and inspection (MIMII dataset) is discussed in \cite{purohit2019mimii}. Normal sounds were recorded for different types of industrial machines (i.e. valves, pumps, fans, and slide rails), and to resemble a real-life scenario, various anomalous sounds were recorded (e.g., contamination, leakage, rotating unbalance, and rail damage). The open source MIMII dataset is to assist the machine-learning and signal-processing community with their development of automated facility maintenance. 

A dataset of household appliances abnormal sound detection \cite{10.1145/3297156.3297186}. \cite{8937164} is another rich acoustic dataset with recorded sound representing operation of miniature machines.

Motor Current data set (MOTOR). Original data set is from \cite{UCRArchive2018}, and contains 420 time series. 21 were chosen including 1 anomalous time series, and each one consists of 1500 real values. Normal time series are the current signals measured from normal operations of a induction motor. The anomalous time series is obtained from a faulty motor.

Power Usage Data (POWER). This data set was obtained from UCR \cite{UCRArchive2018}, and contains 51 time series corresponding to the weekly power consumption

NASA Valve Data (TEK140, TEK160 and TEK170). This data set was also obtained from UCR \cite{UCRArchive2018}. The data values are solenoid current measurements on
a Marotta MPV-41 series valve which is on and off under various test conditions in a laboratory. The normal time series correspond to the data measured during
the normal operations of the valves; the time series data measured during a faulty operation of the valve is considered an anomaly.

Outlier Detection Data Sets (ODDS) \cite{Rayana2016}: This collection provides open access to a number of data sets meant for outlier detection.  In ODDS provides open access to a large collection of outlier detection datasets with ground truth (if available). The focus of ODDS is to provide datasets from different domains and present them under a single platform for the research community. The datasets are arranged based on their types into different tables in the ODDS library.

The Backblaze dataset \cite{backblaze} consists of millions of records pertaining to hardware data for years spanning from 2013 to 2021. Backblaze is a leading data storage provider for enterprises and end-users across the world. The Backblaze dataset is a large-scale dataset with over $228$ million rows of data and a size of approximately $60$ GB. It contains data from four principal hard drive manufacturers- Seagate, Hitachi, Western Digital and Toshiba.The complete dataset consists of several CSV files and each file reports all working hard drives on each day. Each file in the dataset contains several columns, some of which are serial number, model, capacity, failure status and SMART attributes. SMART attributes can be defined as a set of flags which describe the current condition of the hard drive. 

A comprehensive set of open source datasets useful for developing PdM is listed in \cite{2020arXiv200313213L}. These datasets could be leveraged for developing PdM models.



\section{Designing an AI Driven Proactive Customer Care Use-case for Electromechanical Devices} \label{sec_LG}

Inline with the existing technologies \ref{sec:AnomalyPDM}, \ref{sec:fault}, \ref{sec:RL}, \ref{sec:RUL}, detailed case studies based on open source data sets \ref{sec_case_studies}, and an initial exploratory analysis of the data, we adopt a three step approach to derive best business value. Our three step approach has been organized in increasing order of complexity of the use-case and ordered such that the results in the previous stages can be appropriately harnessed to improve the predictive modeling of the next. A detailed description with exemplars are provided next. \\

\noindent \underline{\textbf{Step 1}}: In the first step we adopt a generic fault prediction (GFP) problem. The main idea is to design predictive models to identify/detect a device failure before it happens. The main challenge is to predict such faults within an acceptable time horizon. For this, we plan to use the rich monitoring sensor data (a.k.a superset data) available for specific devices like hard disks  as input; and build anomaly detection models to predict the occurrence of an error `n'-days before it happened (for a given `n'). There are several sub-tasks needed to handle the challenges like, heterogeneity, class imbalance, covariance shift, concept drift, availability of device’s physical model, error codes etc. These include,
\begin{itemize}
    \item Designing appropriate feature mapping (symbolic etc.) to handle data heterogeneity is very important. Most sensor data are categorical and does not have any underlying metric. This causes several issues for effective model building.
    \item Correct/adapt models to covariance shifts or concept drifts resulting due to sensor degradation,
    \item Incorporating a priori knowledge through available `first-principles’ (obtained from device’s physical model). This typically improves the model performance.
    \item Finally designing loss functions which incorporates the ROI quintessential for the use-case success.
\end{itemize}
\textbf{Expected Outcome(s)}: The final outcome from this step will be a Predictive model for device error prediction within a pre-specified time horizon. Such a model can provide several benefits to different stakeholders. For example, a good predictive model can yield significant cost reduction through timely corrections of device conditions and avoiding any reactive maintenance. Further, the output of these predictive models can be tuned to minimize the periodic expert's maintenance visits. Now, rather than periodically scheduling maintenance visits for all deployed devices, we can only target the specific devices likely to fail. This reduces the BU's maintenance costs significantly.   \\

\noindent \underline{\textbf{Step 2}}:  As the next step we adopt a specific fault prediction (SFP) problem. Here, in-addition to identifying a device error we build a predictive model that additionally predicts the specific error type. Similar to the step 1 above, we plan to use the super set data as input. However, different from the above, here we build multi-class predictive models to predict the specific error's likely to occur in the device. There are several challenges in this modeling like, 
\begin{itemize}
    \item Identify the error's more likely to happen or the error codes of specific interest. This task is motivated from the business needs, domain knowledge and can help us build practical models targeted to specific errors of interest. Note that, from a business stand-point, it may not be beneficial to build highly accurate models for errors which rarely occur. Rather, it is more important to predict the errors which are frequent and/or incurs high repair costs.   
   \item Handle the class  imbalance problem or data sparsity per error codes. Imbalance of data for the different error types can bias a model towards the error with more data. To avoid this bias specific cost-sensitive, sampling approaches needs to be designed. 
    \item Another big challenge towards building these multi-class predictive models is the data heterogeneity. Data heterogeneity poses a lot of problems in step 1 too; however these problems increase exponentially for multi class problems. Given that we are trying to estimate multiple phenomenons (unlike just one in step 1), the feature encoding needs to scale well for all the classes.	 
    \item In many cases, the trigger of one error type may lead to the trigger in another error code. Such cases lead to multi-instance, multi-class problems. Handling such problems need special techniques and is non-trivial. 
\end{itemize}
\textbf{Expected Outcome(s)}: Multi-class AI model for predicting device fault types. The outcome of this step can lead practical AI models that can predict the specific errors of interest. This can provide strategic guidance to better planning the maintenance schedules. For example, handling an error which is likely to cause high repair cost demands higher priority compared to ones less costly. Such business metrics can also be built into the predictive models. As another advantage, correctly identifying the type of error could help the experts (serviceman) take appropriate maintenance, repair actions.     

\noindent \underline{\textbf{Step 3}}: Finally the holy grail of predictive maintenance is developing the  Root-Cause-Analysis (RCA) engine for device faults. Note that, identifying errors (in step 1), provides indicators that a device needs attention. Identifying the exact error type provides possible actions to handle the fault. However, in many cases the manifestation of an error code may be due to a ripple effect of an underlying cause. Identifying the right cause can help the serviceman provide the optimal solution. However, pinpointing the exact cause needs several advanced modeling techniques, like causal inference, motif discovery etc. As the final step of our approach, we target building the root cause engine for the device faults. This involves handling a number of challenges which includes (and is not limited to),
\begin{itemize}
    \item Availability of sufficient usage data over time, high granularity of the usage data.
   \item Developing unsupervised learning to understand any clustering effect both in feature space or through motif discovery. 
    \item	Research and develop time-series models both in feature and motif space to correctly capture the root-cause of an error.
    \item Using the RCA and the past action logs to remedy the specific fault types, build an improved recommendation engine on possible actions to prevent such faults. 
\end{itemize}
\textbf{Expected Outcome(s)}: The expected outcome is a RCA engine that can accurately capture the causality model for the device errors. This can yield several benefits to the LG BU's, where in the exact cause for the device fault can be targeted by the repairman, and avoid unnecessary time (and money) in additional diagnostics. This can also provide useful inputs towards improving future device design and manufacturing. 

\section{Lessons learned for enabling a successful PCC use-case} \label{sec_lessons_learnt}
Throughout the survey we have highlighted the several components needed for a PCC use-case. In this section we summarize the lessons learnt for enabling a successful PCC use-case. \\

\noindent \textbf{Business Metric}: The success of any PCC use-case mainly depends on the the availability of the correct metric capturing the business ROI. Blind application of ML models to optimize generic metrics like accuracy, AUC etc., does not guarantee business success. These metrics have to be provided by (or may be collaboratively developed with) the concerned Business Units. For example, in the Anomaly detection exemplar use case discussed in \ref{sec_LG} (\underline{Step 1}) there is a need to understand what proportion of false positives (i.e. false alarms) can be tolerated to correctly identify the errors in a pre-specified number of devices. Alternately, in cases where false positives have minimal impact but the consequences of missing a real fault are very costly it may be prudent to place for more weight on the false negatives in optimization. Such determinations must be made based on business ROI.\\


\noindent \textbf{Data Availability} Availability of quality data is of utmost need for deriving good AI driven business insights. There may be several reasons for a lack of quality data.
\begin{itemize}
    \item \textbf{Type of data collected}. Collecting the right type of data is very important. The collected data should properly align to the use-case's goal. For the exemplar use-case in  \ref{sec_LG} (\underline{Step 1},\underline{Step 2} and \underline{Step 3}), we need the telemetry data (sensor codes, continuous sensor values etc.), sensors collecting a device's environmental variables (temperature, humidity etc.), usage logs (device operation modes, device cycle modes etc.), sensors specific to components of interest (mechanical bearings, compressor etc.), the error codes and finally the correct failure modes. Having quality (non-sparse) data capturing the complete device state through the sensor values is indispensable. 
    
    \item \textbf{Period of data collected}. Data availability over a longer period is very important. Having data > 1 year at least allows to capture the seasonality of device operation (i.e. how the device is operated and behaves in each season). Further, many devices (like Home Appliances, Living Appliances etc.) have a longer life-cycle and tend to fail less often. Availability of data over expected device life-time is highly desirable.
    
    \item \textbf{Error/Failure data availability}. Having substantial error/failure labeled data is very important for building accurate predictive models. Most existing literature \ref{sec_case_studies} targeting a similar use-case as \ref{sec_LG} (\underline{Step 1} or \underline{Step 2} typically expects at least > 1-10 \% of the data to be anomalous (failure data). Building predictive models for extremely rare event failures \cite{chandola2009anomaly} remains mostly unexplored. It is important to remember that for supervised learning algorithms the number of true faults in the training data will usually be the limiting factor in the model performance regardless of the sample size of the negatives in the training set, and even an algorithm with an impressive 99\% symmetric accuracy, will still produce as many false positives as true positives when the class imbalance is at the 1\% level.
    \item \textbf{Data Pre-processing and De-identification}. Any prior data preprocessing, filtering or de-identification needs to be very carefully handled as such operations can corrupt the data statistics and may render the data unusable. For example, in the exemplar use case \ref{sec_LG} \underline{Step 1} certain device identifiers may be de-identified. It is of utmost importance that such a mechanism maintains the uniqueness of the device ids. 
\end{itemize}

\noindent \textbf{Data Readiness} After accessing the required data, there usually remains significant work to do. Some items for consideration include.
\begin{itemize}
    \item \textbf{Data contamination} It is common for data in some applications to be corrupted with mislabeled or missing values. As for many other machine learning tasks, it is essential to make sure the data has been cleaned of outlier and missing values where-ever possible.
    \item \textbf{Feature preparation} Most machine learning applications require all input features to be normalized. For most traditional machine learning algorithms, the engineering of new features from longer scale time series data is usually needed (i.e. average or standard deviations of values of features over certain trailing time periods).
    \item \textbf{Distributed data management} When training or running inference on large datasets it is common for the data to be too large to handle on a single node. Use of a system such as Spark to handle data distribution and distributed model training is often required. The amount of work needed to set up a system like this for the first time should not be underestimated. Therefore, if it is anticipated that data will grow in size to the point where this will be needed in the future it is important to plan ahead for this complication.
\end{itemize}

\noindent \textbf{Model Development Constraints}. For most PCC type use-cases there are several model constraints introduced by BU product specs or during QA process etc. Such constraints need to be imbibed in the model building pipeline before hand. Some typical constraints include (but are not limited to):
\begin{itemize}
    
    \item \textbf{Business needs} The false positive rates, false negative rates, and class imbalance must be carefully considered in context of business requirements. For example, \cite{bib:EKGPaper} presents a model to identify anomalous EKG results with a high accuracy. However, the consequences of being wrong about EKG predictions can have life-or-death consequences. Consequently, they tuned for a recall of identifying normal EKG results of only 30\%, to achieve a precision of 80\%. This would result in saving their cardiologists 30\% time in evaluating patients who do not need help, at the cost of having 20\% of the patients identified as having normal results potentially needing help and not getting it. Acceptable recalls and precisions and recalls must be defined and tuned for, and will usually not be close to the tunings with the highest accuracy.
    \item \textbf{Reduced Model Complexity (for fast inference)}: Most of the time there is limited compute available for integrating any AI into a device. Such compute requirements need to be provided or made available through a BU Specs and Requirements document. A clear understanding of the resource (compute/storage)  requirements is necessary to balance out the optimal model performance while limiting the AI model's resource footprint. 
    \item \textbf{Model Interpretability}: Many use-cases (involving Root Cause Analysis, Action recommendation etc.) require the model predictions to be explainable in human understandable terms. We proposed a new technique that we expect to help with approximate interpretations for a wide variety of models. But no technique is perfect. Typically, the more sophisticated a model is, the more challenging it is to properly interpret. It is recommended to check resulting determinations against the knowledge of domain experts.
    \item \textbf{Hybrid models (Data Driven + Physical models)}: Incorporating the domain knowledge through available physical models or domain expert's recommendations for feature extraction is an important aspect of quality model development. Physics constraints can even be built into the model itself if the physical process is well understood. One strategy to do this is to add the physical constraints directly to the loss function, as described in ~\cite{bib:PhysModels}, which can lead to much faster training, and much better outcomes. It is very important to be sure the physics equation is not an oversimplification of the real world before putting too much weight on an equation in the loss function, however. Close collaboration between the AI modeler and BU experts at all stages of the model building process is recommended. 
    \item \textbf{Advanced model selection} For any application, there are dozens of machine learning algorithms available that can be chosen, and it is hard to know in advance which is best for the job. A useful tool for selecting the ideal algorithm which is worth considering is presented in~\cite{bib:TSFreshAlgoSelect}, which trains an XGBoost model to predict the best model to use for a particular anomaly detection task. In it, a dataset of example time series is paired with the ideal algorithm choice, passed through the tsfresh package to construct time series features which are then filtered down by a genetic algorithm, and then used to train a model to predict the best model for the particular Anomaly detection task.  
\end{itemize}

In all, the above aspects are very important for enabling a successful AI driven PCC use-case, and should be carefully considered.

\section{Conclusions} \label{conclusions}
This work provides a brief survey on Proactive Customer Care. The survey identifies the different types of maintenance services provided by OEMs and highlights the importance of Predictive Maintenance (a.k.a Proactive customer care). It addresses the recent technologies in machine learning algorithm and system design to handle a typical AI driven Predictive Maintenance use-case. We cover the existing external use-cases and provide a list of open source data. Finally, we also provide the lessons learned towards enabling a successful PCC use-case.

\section{Acknowledgement}
We would like to thank our colleague at LG Electronics Advanced AI team, Samarth Tripathy, for assistance with web deployment of the live demo for the Microsoft Azure dataset. We also acknowledge the Google Cloud Platform (GCP) team for providing us with the computing resources for the analyzing the Blackblaze dataset.

\bibliographystyle{unsrt}
\bibliography{PdM}



\end{document}